\documentclass[journal]{IEEEtran}
\usepackage{pifont} 
\usepackage{amssymb} 
\usepackage{caption}
\usepackage{hyperref}
\usepackage{amsmath}
\usepackage{xcolor}
\usepackage{amssymb}  
\definecolor{mblue}{RGB}{50, 100, 200}

\ifCLASSINFOpdf
  \usepackage[pdftex]{graphicx}

\else
\fi
\begin{document}
%
\title{TSGAN: An Optical-to-SAR Dual Conditional GAN for Optical based SAR Temporal Shifting}
%
%
%

\author{Moien~Rangzan,
        Sara~Attarchi*,
        Richard Gloaguen,
        Seyed~Kazem~Alavipanah

\thanks{M. Rangzan, S. Attarchi, and S.K. Alavipanah are with the Department of Remote Sensing and GIS, Faculty of Geography, University of Tehran, Tehran, Iran.}
\thanks{R. Gloaguen is with the Department of Exploration technology, Helmholtz-Institute Freiberg for Resource Technology, Freiberg, Germany.}}

%
%

\markboth{Preprint, \today}%
{Shell \MakeLowercase{\textit{et al.}}: TS‑GAN: An Optical-to-SAR Dual Conditional GAN for Sentinel‑1 Temporal Shifting using
Optical Imagary}
%



\maketitle

\begin{abstract}
In contrast to the well-investigated field of SAR-to-Optical translation, this study explores the lesser-investigated domain of Optical-to-SAR translation, a challenging field due to the ill-posed nature of this translation. The complexity arises as a single optical data can have multiple SAR representations based on the SAR viewing geometry. We propose a novel approach, termed SAR Temporal Shifting, which inputs an optical data from the desired timestamp along with a SAR data from a different temporal point but with a consistent viewing geometry as the expected SAR data, both complemented with a change map of optical data during the intervening period. This model modifies the SAR data based on the changes observed in optical data to generate the SAR data for the desired timestamp. Our model, a dual conditional Generative Adversarial Network (GAN), named Temporal Shifting GAN (TSGAN),  incorporates a siamese encoder in both the Generator and the Discriminator. To prevent the model from overfitting on the input SAR data, we employed a change weighted loss function.
Our approach surpasses traditional translation methods by eliminating the GAN's fiction phenomenon, particularly in unchanged regions, resulting in higher SSIM and PSNR in these areas. Additionally, modifications to the Pix2Pix architecture and the inclusion of attention mechanisms have enhanced the model's performance on all regions of the data. This research paves the way for leveraging legacy optical datasets, the most abundant and longstanding source of Earth imagery data, extending their use to SAR domains and temporal analyses. To foster further research, we provide the code, datasets used in our study, and a framework for generating paired SAR-Optical datasets for new regions of interest. These resources are available on GitHub at \href{https://github.com/moienr/TemporalGAN}{github.com/moienr/TemporalGAN}

\end{abstract}




%
\IEEEpeerreviewmaketitle

\section{Introduction}

\IEEEPARstart{G}{e}nerative Adversarial Networks (GANs) have played an increasingly significant role in the field of remote sensing. Among their diverse applications, i.e., semantic segmentation \cite{FAN2023OilSemantic},  super-resolution \cite{ZHU2023superres}, text-to-image generation \cite{zhao2022texttoimage} there has been a notable emphasis on data translation, particularly in the context of image-to-image translation\cite{JOZDANI2022ganreview}. This technique involves creating mapping functions that connect input and output data \cite{zhu2017cyclegan}. This method finds versatile applications, encompassing domain adaptation (DA) and the conversion of diverse remote sensing data sources. Its goal is to enhance model performance in downstream tasks and/or improve the interpretability of data\cite{JOZDANI2022ganreview}.

Translation between Synthetic Aperture Radar (SAR) and Optical data is a significant domain where image-to-image translation is increasingly applied in remote sensing\cite{JOZDANI2022ganreview}.

The SAR-to-Optical (SAR2Opt) translation serves two primary objectives: first, it aims to enhance the interpretability of SAR data; and second, it leverages the all-weather, night-day capabilities of SAR instruments, rendering them invaluable sources for cloud removal tasks\cite{chen2021attentive,li2020thincloud,xiong2021noncloud} or as an alternative to optical data when it is unavailable due to thick smoke and aerosol layers in the atmosphere \cite{zhao2022comprehensivegan}.

Despite significant advancements in SAR2Opt translation, driven by the advent of GANs, there remains a notable gap in the literature when it comes to a comprehensive analysis of the reverse challenge, the Optical-to-SAR Translation (Opt2SAR). This gap is attributed to the inherently challenging nature of the problem, particularly in dealing with the dissimilarity between two SAR data within a single Region of Interest (ROI) when the viewing geometries differ \cite{fu2021reciprocal}. Nonetheless, achieving a robust model on this side of translation can pave the way for translating legacy optical datasets that have been made prior to a SAR mission. This advancement can also improve downstream tasks, such as the process of detecting changes between heterogeneous SAR and optical images \cite{zhao2022comprehensivegan}, and contribute to the development of super spatial-temporal resolution models for SAR images \cite{JOZDANI2022ganreview}. This is achievable by supplementing low spatial/temporal resolution SAR data with high spatial/temporal resolution optical data,i.e., using optical time series to fill in the gaps in the SAR time series or leveraging it to increase SAR's spatial resolution.

A recurring challenge in both SAR2Opt and Opt2SAR translation literature is the "Fiction" phenomenon, which occurs when the reference data lacks sufficient information. This deficiency leads the model to supplement its own data, often deviating from the actual ground truth, to generate the translated data. The Fiction issue is particularly evident in two areas of a SAR2Opt task. Firstly, the GAN model struggles to restore the actual spectral diversity of the ground truth data, leading to a low fidelity in spectra in the translated optical data \cite{doi2020sar2opticcolorregion}. Secondly, the model often fails to retain texture and fine-grained borders in the translated data, as SAR and optical data have fundamentally different structures, resulting in an inaccuracy of detail \cite{YANG2022improvedcgancolorloss, reyes2019sar2optic}.

Regardless of whether it is a SAR2Opt or Opt2SAR task, most existing research in this domain has relied on mono-temporal SAR-Optical datasets. It means that the algorithms are set to learn the relationship between SAR and optical domains without additional information. This would require that SAR and optical data provide a similar discrimination of surface targets. However, this is not the case, as surfaces with the same spectral composition can have different backscattering properties and vice versa. To counter this problem, Xiong et al. \cite{xiong2021cloudmultitemp}, in their SAR2Opt task, concatenated the SAR and optical data of a previous timestamp in order to keep the surface details and save textural and spectral details in the generated optical data. This is close to the approach taken by He et al. \cite{he2021stanfordsuperres}, where in their super-resolution task, they incorporated high-resolution data with different timestamps alongside the low-resolution data of the desired timestamp to generate high-resolution data.

Nonetheless, incorporating data from the same domain as the desired output introduces a bias; If the changes between the input data of the desired domain and the output are small throughout the whole dataset, it will result in an imbalanced dataset, overfitting the model on the input, where the model learns to merely copy the input data as the output. At the same time, the metrics still show favorable results \cite{JohnsonKhoshgoftaar2019}. Our research addresses this challenge by offering a comprehensive solution and evaluation.

To address the existing gaps in Opt2SAR research and tackle the issue of inaccuracies in image translation, i.e., fiction phenomenon, we have developed a novel strategy we call "SAR Temporal Shifting." This approach redefines the Opt2SAR translation problem by considering how an input SAR image should be modified in response to changes observed in optical data over time in order to align with a new SAR image from a desired timestamp. To implement this concept, we designed an attention-based GAN architecture, incorporated a change-weighted loss function to minimize overfitting to input data, and introduced new performance metrics. Additionally, we created a bi-temporal dataset with consistent SAR viewing angles to facilitate this study. The key contributions of our work can be summarized as follows:

\begin{enumerate}
    \item \textbf{Introduction of a novel attention-based siamese encoder UNET Architecture GAN:} This architecture is designed to take optical data from a specified timestamp as input and SAR data from a prior timestamp, both complemented with the changes in the optical data at both timestamps. Its purpose is to establish the foundational geometry of the resultant SAR data using the old SAR data and then train the model to adapt the input SAR data to match the optical data changes, thus facilitating the transition to a new timestamp.

    \item \textbf{Design of an automatic GEE Paired Sentinel-1 and Sentinel-2 dataset downloader framework: }This workflow autonomously identifies the most suitable Sentinel-1 SAR orbit according to specified criteria. Subsequently, it retrieves cloud-free Sentinel-2 optical data, pairs it with despeckled SAR data, and divides them into patches, thereby constructing a meticulously paired dataset. Moreover, this framework is deliberately engineered for effortless customization to accommodate diverse geographical regions, enhancing its overall utility and facilitating its seamless integration into forthcoming research pursuits.

    \item \textbf{Leveraging a change-weighted L1 loss function: }To account for the inherent imbalance in the altered areas within an urban setting over a short time span, we propose a specialized loss function. This loss function is crucial in preventing the model from merely replicating old SAR data as the new one by assigning higher weights to the changed regions.

    \item  \textbf{Implementation of change-weighted metrics:} In the evaluation phase, we introduce metrics that distinguish between the model's performance in areas that have undergone changes and those that remain unchanged. This distinction enables a more nuanced assessment of model effectiveness.
\end{enumerate}

\section{Related Work}

\subsection{Generative Adversarial Networks}

Generative Adversarial Networks (GANs) operate on the concept of a min-max game, where both the generator and the discriminator are iteratively refined to reach a Nash equilibrium. The goal is to make the distribution of generated data closely resemble that of real data. In this process, the generator produces synthetic data from noise, typically drawn from a uniform or Gaussian distribution. Meanwhile, the discriminator's role is to discern between outputs from the generator and actual data. The aim of the generator is to better mimic real data, and for the discriminator, it's to enhance its ability to distinguish real data from synthetic ones. While GANs are generally effective, they are unsupervised networks, which can sometimes limit the scope and control of the generated outputs \cite{YANG2022improvedcgancolorloss}. To address this limitation, conditional GANs (CGAN) are introduced \cite{mirza2014cgan}. CGAN incorporates additional conditions, such as labels, data, or text, to constrain the types of generated data. The loss function for CGAN is defined as follows:

\begin{equation}
\begin{aligned}
\min _G \max _D \mathcal{L}_{c G A N s}(G, D)= & E[\log (D(y, x))] \\
& +E[\log (1-D(y, G(x)))]
\end{aligned}
\end{equation}

In this expression, $x$ represents actual data, $y$ indicates the conditional parameters, and $G(x)$ is the output from the generator. The term $D(y, x)$ is the discriminator's output when given real data, while $D(y, G(x))$ is how the discriminator responds to the generator's synthetic data.

Image-to-image translation GANs are CGANs in which the imposed condition is the input image. Based on the optimization method and the arrangement of source–target datasets, current GAN-based methods for image-to-image translation can be categorized into two main groups: paired and unpaired methods \cite{zhao2022comprehensivegan}.

The paired methods need a dataset with matching source and target data, i.e., for each source data, there must be a target data. A landmark for paired methods is the Pix2Pix\cite{isola2017pix2pix}architecture. Its generator follows a U-net architecture, while the discriminator employs a patchGAN architecture to model the image as a Markov random field\cite{wang2019cycle}.

In contrast to the paired translation, unpaired methods use two unpaired datasets from both domains. A standard method for unpaired methods is CycleGAN \cite{zhu2017cyclegan}, which incorporates a combination of cycle consistency loss and identity loss in the GAN loss function.

\subsection{Optical-SAR translation}

The literature on optical to SAR translation can be categorized into two sections: the body of research that focuses on the translation aspect of the problem and the research that aims to harness this method for practical applications. The following subsection contains a review of these categories.

\subsubsection{Translation methods}

   The comparative evaluation by Zhao et al. \cite{zhao2022comprehensivegan}, was performed on various data-to-data translation methods for SAR2Opt translation, including BicycleGAN \cite{zhu2017BicycleGAN}, CycleGAN \cite{zhu2017cyclegan}, CUT \cite{radford2015cut}, MUNIT \cite{huang2018munit}, NICE-GAN \cite{chen2020reusingnicegan}, and Attn-CycleGAN \cite{lin2021attentionattncyclegan}. The evaluation was conducted on the SEN1-2 dataset \cite{schmitt2018sen12}, a paired dataset of Sentinel-1 and Sentinel-2 data captured in 2016 and 2017. The dataset was divided into four categories based on seasons, i.e., Winter, Fall, Summer, and Spring. It covers a vast geographical area whereas the sample points were biased towards urban areas. Furthermore, in their study, Zhao et al. \cite{zhao2022comprehensivegan} also introduced a novel dataset called SAR2Opt, which consists of paired data from TerraSAR-X and Google Earth. The evaluation results indicated that Pix2Pix and CycleGAN outperformed other models in terms of peak signal-to-noise ratio (PSNR) and structural similarity index (SSIM). Pix2Pix exhibited the highest PSNR scores across all datasets and the best SSIM in the Winter, Fall, and SAR2Opt datasets. On the other hand, CycleGAN achieved superior SSIM scores in the Spring and Summer datasets. The authors attributed this distinction to the complexity of land surface characteristics, for which CycleGAN's unpaired approach was better suited. This compelling performance of Pix2Pix has led us to select it as the base model for our study, as discussed in the methodology section of this paper.

Reyes et al. \cite{reyes2019sar2optic} selected CycleGAN as their base model for SAR2Opt translation, with a primary focus on result interpretability. The authors acknowledge the fundamental limits of SAR2Opt translation that cannot be compensated for, suggesting that learning the transition from a single-channel SAR data to a multi-channel optical data is an inherently challenging, ill-posed problem similar to colorizing gray-scale data in classical computer vision. They emphasize that the diversity of surface parameters, such as variability and correlation length, relative permittivity, or geometry, which contribute to unique wavelength- and temperature-dependent signal responses, makes the work more challenging.
Reyes et al. also identified issues related to the Fiction phenomenon in the translation process. They demonstrated that, in urban areas, while models could be successful in transforming regions with corner reflection into blocks of buildings, the shape and number of buildings in the resulting optical data were different from those in the original data. Another challenge was the separation of agricultural fields, which is immediately discernible in optical data but fades in SAR data and thus its optical translation.

A reciprocal Optical-SAR translation study was done by Fu et al. \cite{fu2021reciprocal}, where they introduced a modified Pix2Pix GAN architecture that uses multi-scale cascaded residual connections, i.e., CRAN. They tested their model on both satellite and aerial datasets. They concluded that when using full PolSAR data as input, the results were more favorable compared to using single polarisation data. They attributed the discrepancy to the fact that some objects are not visible in single-pol SAR data. Similar to Reyes et al., they found that, while land covers like waters and vegetation areas were easily reconstructed, the model struggled to reconstruct built-up areas, resulting in building cubes with edges that are not well-aligned. Furthermore, for tall buildings, the structures were smeared, which they attributed to model confusion due to variable viewing angles.

Intertwined with the discourse presented by Reyes et al., we recognize that the task of translating optical data into SAR data is not a straightforward one-to-one problem due to artifacts introduced by surface topography and the varying viewing geometry of the SAR instrument. Our study provides a comprehensive discussion of the challenges as well as our approach to overcoming these challenges.

\subsubsection{Applications of optical-SAR translation}

Li et al. present the Deep Translation network for Change Detection (DTCDN). This network leverages the innovative Nice-GAN architecture as the translation model and incorporates a customized U-Net++\cite{Zhou208unetpp} architecture with a multi-scale loss function for change detection. In essence, their approach involves the transformation of one data, either SAR or optical, to the domain of the other data. Subsequently, both translated and original data of another time are fed into the U-Net++ network, which is then trained in a supervised manner to detect changes. In a thorough evaluation of four different datasets, they found that the effectiveness of translation varied: in some cases, translating optical to SAR was more successful, while in others, SAR to optical yielded better change detection results\cite{Li2021deeptranslationchangedetection}.

In a recent study by Hu et al., a comprehensive dataset comprising Sentinel-1 and Sentinel-2 data was employed to successfully showcase the potential of GANs for SAR2Opt translation in the context of over 304 wildfire events. The researchers capitalized on the unique capability of SAR data to capture surface details even in the presence of dense smoke, which tends to obscure optical wavelengths and renders optical data ineffective.
To accomplish this, the authors utilized a ResNet-based Pix2Pix architecture for translating SAR data, both before and after the wildfire events, to generate burn severity classification maps and optical burnt indices using all possible combinations of real and fake optical data from the two timestamps. Notably, the study's findings revealed that the generated indices exhibited a stronger correlation with the actual indices compared to the original SAR-based indices. Moreover, this approach surpassed SAR-based indices in terms of accurately mapping the extent of burnt areas\cite{hu2023wildfire}.

\section{Bitemporal dataset}
In our study, we recognized the importance of having a consistent multi-temporal dataset that could be used to effectively test and refine our methodology. To address this need, we developed two paired datasets consisting of Sentinel-1 GRD and Sentinel-2 data from the years 2019 and 2021. The Sentinel-1 data exclusively featured the VV polarization, while the Sentinel-2 data comprised RGB and NIR bands at a spatial resolution of 10m, along with the SWIR-1 and SWIR-2 bands at 20m. The Sentinel-2 data were obtained from the Level-2A Collection, which includes atmospherically corrected surface reflectance (SR) data derived from the associated Level-1C products.

\begin{figure*}[t]
\centering
\includegraphics[width=7in]{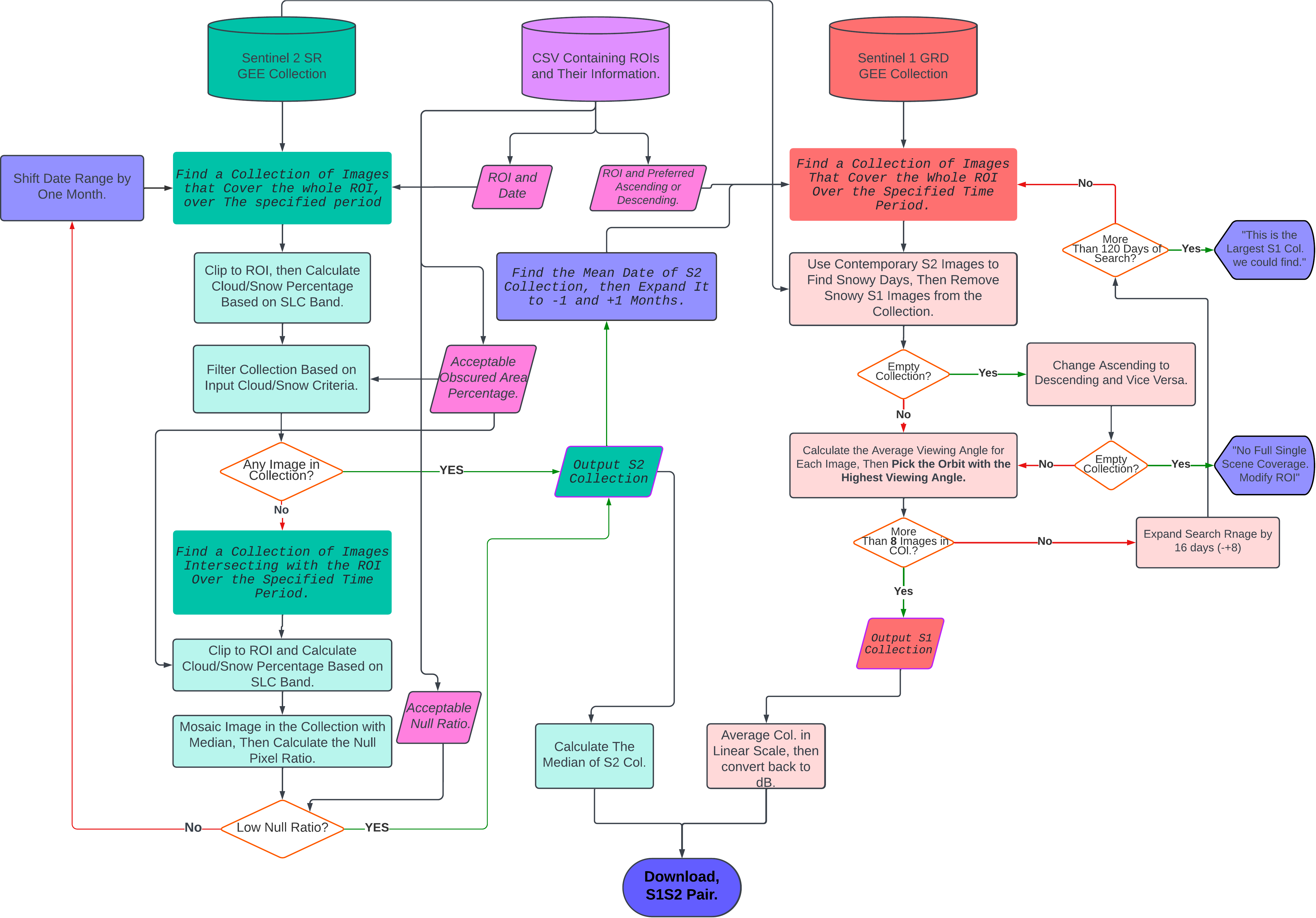}
\caption{ \label{fig:dataset Flowchart} Flowchart of Automated GEE Paired S1S2 dataset Downloader}
\end{figure*}

To ensure reproducibility, we developed a semi-automated workflow using the Google Earth Engine (GEE)\cite{gorelick2017gee} Python API and the GeeMap\cite{Wu2020geemap} Library. Our workflow, available on GitHub, allows the user to create a paired Sentinel-1 and Sentinel-2 dataset by downloading and patching data for a given set of ROIs and dates. The process requires only a few parameters to be tuned. We will delve into the specific details of this workflow and the resulting dataset in subsequent sections. Figure\ref{fig:dataset Flowchart} provides a high-level overview of the workflow.

\subsection{Sentinel-2 data}
The process of acquiring Sentinel-2 data through our workflow is relatively straightforward. Given the geographical location and corresponding date of ROI, the process locates a GEE data collection of that area. The cloud cover of the ROI is calculated using the Scene Classification (SLC) band in the Level-2A Collection\cite{raiyani2021scl}. If a scene or data collection is not found with an acceptable level of cloud cover, the process recursively shifts the date by one month until a suitable collection is located.

Once a collection is found, we select either the data with the lowest cloud cover or the median of the data collection, depending on the number of data in the collection. This approach ensures that we acquire the most suitable Sentinel-2 data available for each ROI and date.

\begin{figure}[!t]
\centering
\includegraphics[width=3in]{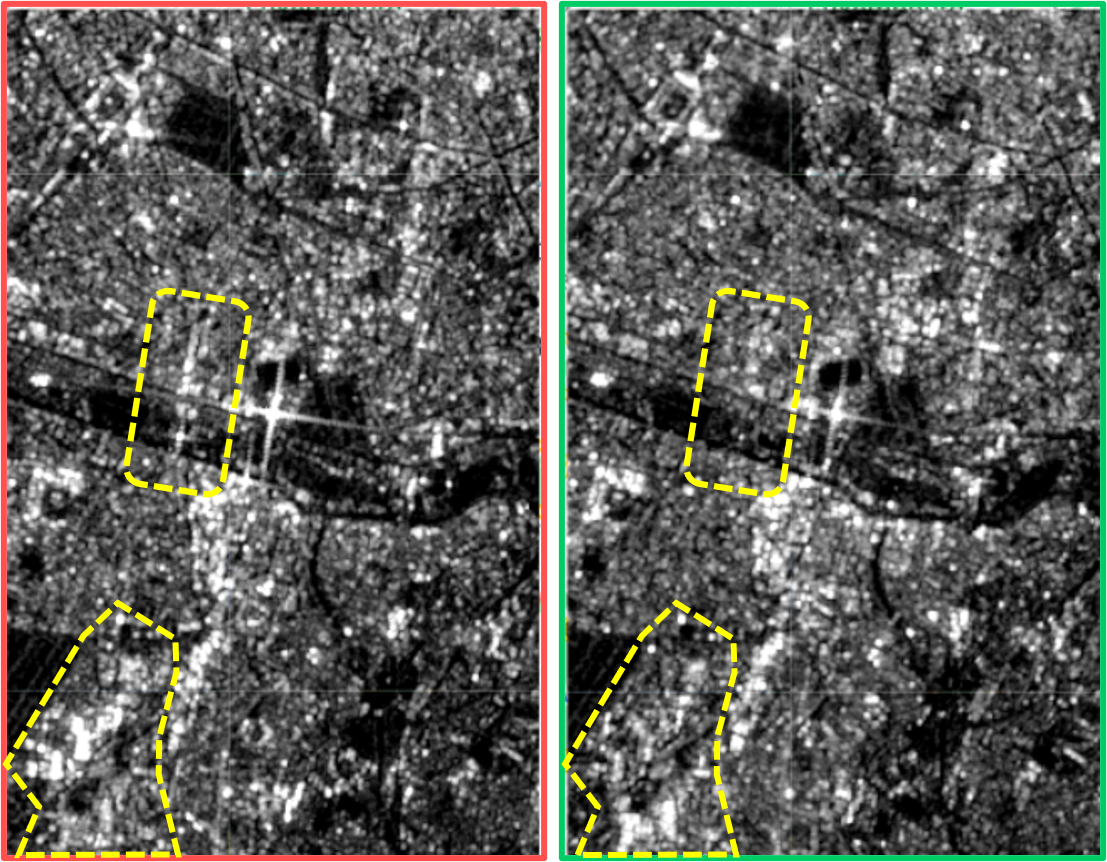}
\caption{Difference in the Sentinel-1 VV ascending data due to different incident angles - Bercy, Paris, France. (48.82N, 2.33E)}
\label{fig:diff_or}
\end{figure}

\subsection{Sentinel-1 Data}
As previously stated in the introduction, the Opt2SAR translation presents an ill-posed problem where, based on the SAR instrument's viewing geometry, there are multiple solutions for this translation.

Figure \ref{fig:diff_or} provides a visual representation of this phenomenon, depicting two ascending data captured over Paris, with different incident angles (The incident angle is the angle between the incoming radar beam and a vector perpendicular to the target \cite{duguay202150yearlake}). It can be observed that the backscatter values, the shape of the building blocks, and SAR artifacts are different in these two data. 
In order to address the issue of indeterminate solutions, a set of rules on the selection of the region of interest, orbit number, and despeckling was established to constrain the possible outcomes, aiding the GAN model in generating a unified answer for the translation. The following sections will provide a more detailed explanation of how each of these rules has been implemented.

\subsubsection{Regions of interest}
SAR data is subject to significant influences from the Earth’s relief \cite{woodhouse2017introduction}. In regions where the relief exhibits high variance, such as hills and valleys, SAR data can be affected by foreshortening, layover, and occlusion. To avoid these issues, we chose to limit the Regions of Interest (ROIs) to urban areas, which are generally flat and exhibit low relief variance. The ROIs were selected using the Onera satellite change detection (OSCD) dataset\cite{daudt2018oscd}, which includes 24 urban areas from around the world. Of these, 14 were designated for training, and the remaining ten were reserved for testing. The OSCD dataset was originally developed for change detection analysis of Sentinel-2 data.

\begin{figure*}[t]
\centering
\includegraphics[width=7in]{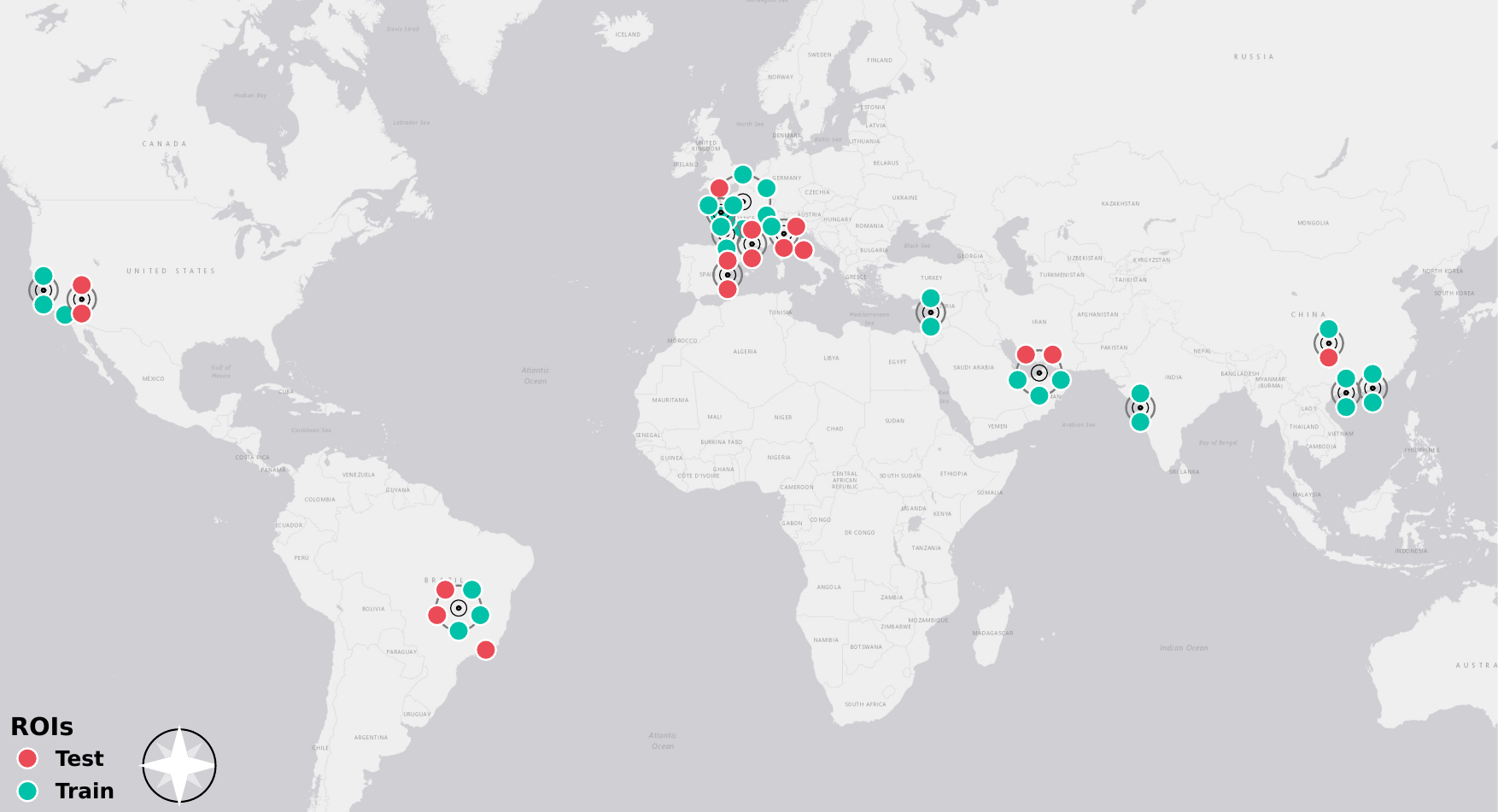}
\caption{ \label{fig:s1s2map} Spatial Distribution of dataset}
\end{figure*}

To increase the size of the OSCD dataset for training the generative adversarial network (GAN), we employed two approaches. Firstly, we expanded the regions to encompass entire urban areas since some of the original OSCD ROIs represented only a small portion of a city. We carefully selected new ROIs that did not contain hills or valleys in order to avoid the issues associated with geometric artifacts in SAR data since these topographic variances, while barely visible in the optical images, can heavily impact the SAR image, resulting in confusion of the model. The second approach involved the addition of adjacent cities to the existing dataset. We made the assumption that neighboring cities would exhibit similar architectural styles and use of materials, thus making them suitable for inclusion in the expanded dataset. By employing both of these approaches, we were able to increase the number of ROIs in the dataset from 24 to 46. These ROIs were split into two sets: a training set of 30 ROIs and a testing set of 16 ROIs. The expanded dataset enabled us to train the GAN on a more extensive and diverse range of urban areas, which we anticipate will enhance the model’s ability to generate realistic translated SAR data. For reference, Figure \ref{fig:s1s2map} provides a visual map pinpointing the geographical locations of the ROIs utilized in this extended dataset.

\subsubsection{Orbit number}
As previously noted, multiple orbits of the Sentinel-1 instrument can capture the same ROI, and these orbits can have a significant impact on the resulting SAR data. To enhance the homogeneity of our dataset across different spatial locations and over time, we developed a strategy to average the SAR incident angle band, which is provided as a Sentinel-1 band in GEE, over the ROI. Specifically, when multiple orbits are available for a given ROI, our workflow selects the orbit with the highest average incident angle over the given ROI. This approach aims to reduce the variability in incident angles across different ROIs and mitigate distortions such as foreshortening and layover\cite{woodhouse2017introduction}. Moreover, we ensured that for each ROI, selected orbits were consistent in both 2019 and 2021 acquired images. By implementing these measures, we aimed to increase the generalizability and robustness of our dataset.

\subsubsection{Despeckling}

In the field of SAR2Opt data translation, the task of despeckling has often been left to GAN models. However, when performing the reverse task of Opt2SAR data translation, failing to address speckle in the preprocessing steps can lead the model towards attempting to add speckle-like noise to the output. This results in poorer performance due to the inherent randomness of speckle \footnote{While speckle is practically random, strictly speaking, it is a deterministic and repeatable phenomenon under identical conditions.}. To mitigate this problem, we have developed a workflow that involves identifying the mean date within a Sentinel-2 data collection and expanding the timeframe to 2-3 months (or up to four months in a few extreme cases) to acquire between 8 to 10 Sentinel-1 data. These data are then converted into a linear scale, averaged, and the resulting despeckled data is returned in a logarithmic scale. By consistently selecting a similar number of data, this method ensures that each data in the dataset undergoes the same level of despeckling, creating a more generalizable dataset for the model. It is worth mentioning that we preferred to take a temporal approach for speckle filtering since we expect the urban environment to show no or minimal changes during the averaging period, thus avoiding the blurriness caused by mono-temporal spatial filters.

\subsubsection{Patching}

In order to input each  ROI into our network, we divided them into patches with a size of 256x256. To minimize the number of leftover pixels in each data, we employed an adaptive approach to find the optimal vertical and horizontal overlap.

Despite our efforts to curate a dataset with consistent incident angles, uniform relief, and standardized despeckling, we recognized that these measures alone might not fully address the complex challenges inherent in Opt2SAR translation. The inherent intricacies of this task, stemming from the nature of SAR and optical data, suggest that a more nuanced approach in our methodology is required. This leads us to explore innovative techniques in the subsequent methodology section.

\section{Methodology}

To address the issue of uncertainty in translating Opt2SAR data, we adopted a solution that involved using an older Sentinel-1 data ($S_1T_1$) as input, along with the current Sentinel-2 data ($S_2T_2$) to generate a Sentinel-1 data for the current timeframe ($S_1T_2$). By doing so, we transformed the problem of “translating optical data to SAR data” into the question of “how SAR data would change based on the changes in optical data” or "SAR  temporal shifting," a term borrowed from video generation literature\cite{munoz2020temporal} as our method was inspired by next frame generation deep learning papers, where the aim is to predict the next frame of a video, based on previous frames time series or induced conditions\cite{donahue2019labelconditioned}.
However, this innovative approach also introduces a formidable challenge, which we have termed the "curse of copy-and-pasting."

The temporal resolution of two years may not be sufficient to capture significant changes in urban areas, as urbanization is a gradual process that occurs over longer time periods\cite{urbangrowthrate}. This limitation can lead to several areas in the dataset experiencing minimal or no changes in the span of two years. This causes a class imbalance of changed and unchanged classes, which, in turn, can overfit the model to become vulnerable to the copy-and-paste problem, where it merely reproduces the $S_1T_1$ input as the $S_1T_2$ output, with the model's loss remaining low. Moreover, the metrics are likely to demonstrate favorable results\cite{JohnsonKhoshgoftaar2019}. 

In the subsequent sections, we will first elaborate on the model’s architecture, how we redesigned Pix2Pix architecture to fit our specific problem, and how we implemented attention mechanisms to further improve it. Second, we will discuss how we used a new cost function to mitigate the problem of copy-and-pasting. Finally, we will introduce new weighted metrics to evaluate our model's performance on both changed and unchanged regions separately.

\subsection{Base GAN architecture}

In defining the architecture of our model, we initially adopted the Pix2Pix framework as our foundation. However, we made specific modifications to both the generator and the discriminator to fit the requirements of our task.

\subsubsection{Generator}

In our base model, we leveraged the Pix2Pix architecture but introduced a significant adaptation by incorporating two encoders within the U-Net architecture. These encoders were dedicated to encoding the SAR and optical data streams separately. Having separate branches for SAR and optical data is a strategy that showed promising results in fusing SAR and optical data for change detection\cite{hafner2022urban}. At the bottleneck layer, the encoded data from both streams were concatenated. Subsequently, the concatenated data underwent upsampling through the decoder. Each step of both encoders was connected to the upsampling decoder through a skip connection. This structural modification, reminiscent of Pix2Pix, is referred to as DE-Pix2Pix.

In order to further sharpen the model's output and circumvent the issue of blurring, which was even noticeable in areas with minimal changes— where we expect the model to do the task of copying and pasting— we opted to replace the initial downsampling layer with a 1:1 convolution layer, effectively eliminating downsampling. This decision was made as our investigation revealed that this blurriness was primarily attributable to the first skip connection, positioned after a downsampling layer. This configuration forced the last upsampling layer of the Pix2Pix model to generate a 256x256 feature from a 128x128 feature and neglecting lower-level feature maps in the SAR branch, resulting in unnecessary data distortion.

In this new setting, we chose a 3x3 kernel size for the SAR encoder initial layer, while a 5x5 kernel size was employed for the optical data layer. The choice of the larger kernel size for optical data was necessitated by the need to provide a wider field of view. This decision was taken by the understanding that the correspondence between Optical and SAR data is not strictly pixel-to-pixel, as neighboring pixels can exert influence on one another (for instance, due to SAR artifacts).

The remainder of the network retained its original structure, with the exception of the last upsampling layer, which was replaced with a 1:1 convolution layer featuring a 3x3 kernel size.

\subsubsection{Discriminator}

\begin{figure}[!t]
\centering
\includegraphics[width=3in]{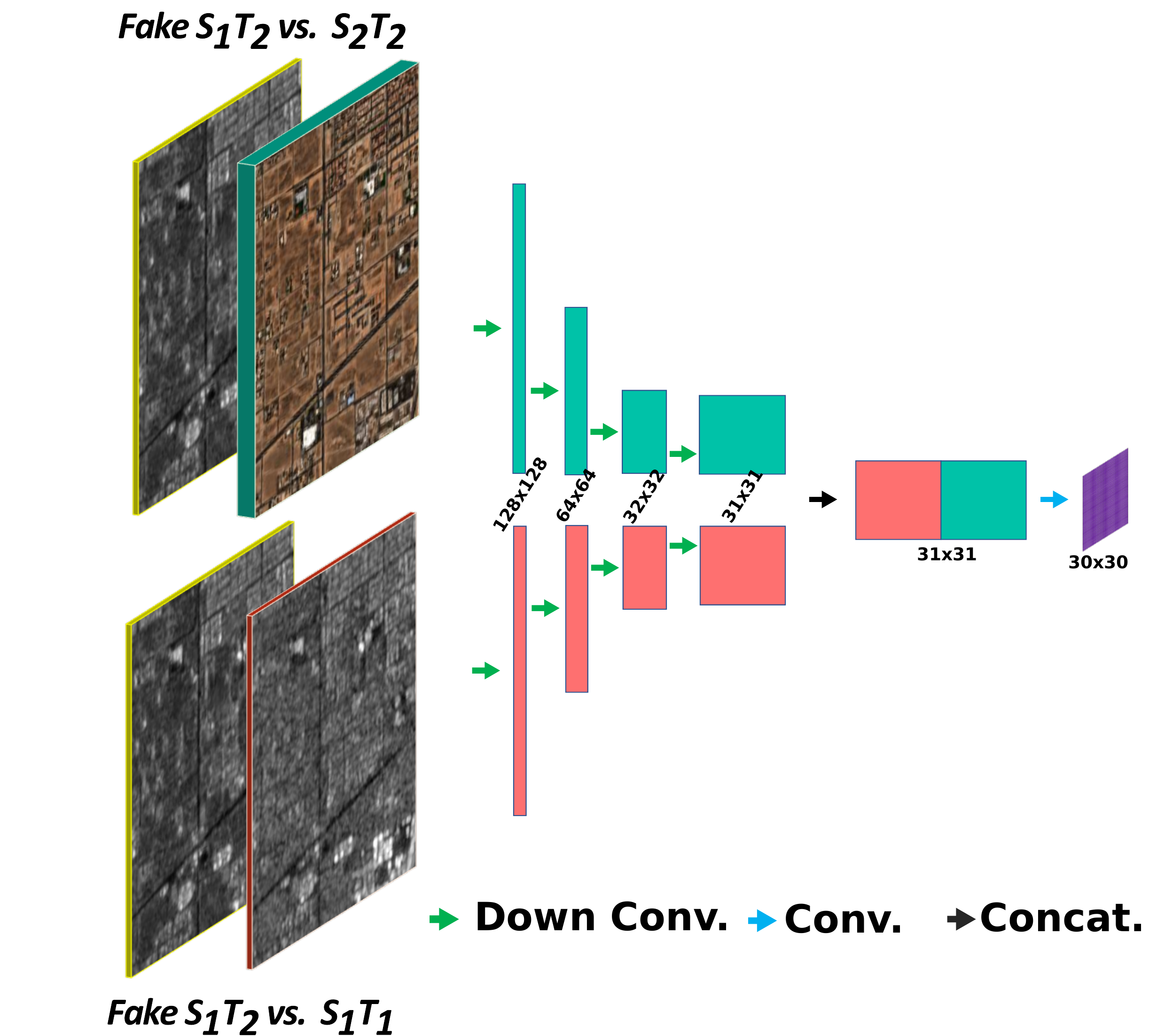}
\caption{Architecture of the Discriminator}
\label{fig:discriminator_architecture}
\end{figure}

The discriminator of our model follows the patch-GAN architecture utilized in Pix2Pix. However, to adapt the model for a dual input architecture, we made adjustments by incorporating two parallel encoders. The input to each encoder is the generated $S_1T_2$, stacked over each of the inputs. After downsampling, both streams are fused into a patch of 30x30 output, the same as the original model. Figure\ref{fig:discriminator_architecture} provides an illustration of the discriminator architecture.

\begin{figure*}[t]
\centering
\includegraphics[width=7in]{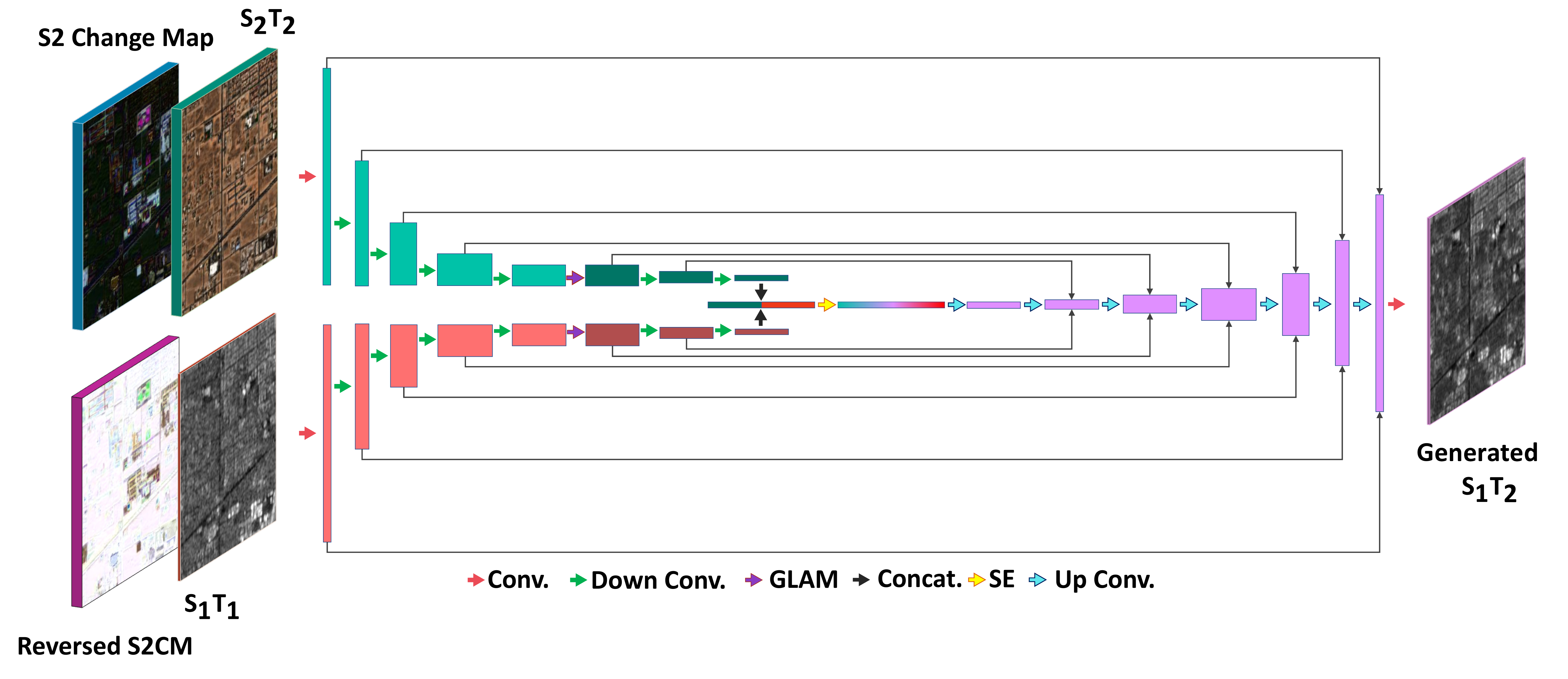}
\caption{Architecture of the Generator}
\label{fig:generator_architecture}
\end{figure*}

\subsection{Attention based architecture}

With the surge in popularity of transformers in various data science applications \cite{vaswani2017attention}, notable advancements have emerged in the utilization of vision transformers \cite{Dosovitvisiontransformer,tokenvisiontransformer} within the domain of remote sensing, particularly in the realm of SAR data analysis \cite{boutosvitsar,chenvitcd,pangvitcdsar}. 

Numerous scholarly contributions have endeavored to enhance data autoencoders and, notably, the U-Net architecture through the integration of attention mechanisms \cite{oktay2018attentionunet,khanh2020enhanceunetatteniton,zhao2020scanet}. Furthermore, attention mechanisms have demonstrated their efficacy in improving the performance of GANs by passing the target information through a weighted map\cite{lin2021attentionattncyclegan}, and their applicability has extended to the realm of remote sensing as well \cite{chen2021attentive}.

Building upon this foundation and to further bolster the model's performance, we harnessed two distinct attention mechanisms.

First, we utilized channel attention based on the well-known squeeze and excitation(SE) paper\cite{HuShenSun2018SqueezeExcitation} prior to the bottleneck fusion layer. This was inspired by Rangzan et al.'s research\cite {rangzan2022stripe} demonstrating that when dealing with multi-modal data, a fully connected layer at the fusion level of U-Net can enhance its performance for segmentation tasks. SE allows the network to determine the appropriate weighting for the amalgamation of SAR and Optical encoded data, which is then passed to the decoder segment of the network.

Furthermore, the importance of data in each stream can spatially vary. For example, the model might need to utilize the data of unchanged areas from the Sentinel-1 data while reconstructing the data of changed areas from the optical Sentinel-2 data. to help the model to better focus on different parts of each Sentinel-1 or Sentinel-2 stream, we implemented the Global-Local Attention module (GLAM) \cite{SongHanAvrithis2023GLAM} at the downstream features of both encoder streams in the generator model. 

GLAM combines both global and local attention mechanisms. In the context of GLAM, local attention involves capturing interactions among nearby positions and channels. This is achieved using techniques like pooling (similar to Convolutional Block Attention Module (CBAM)\cite{cbamwoo}) for dimension reduction, as well as convolution kernels for channel and attention map derivation. However, it is important to note that this local attention approach is spatially limited in its ability to establish relationships among neighboring features due to the constraints imposed by the size of the convolution kernel. 

On the other hand, the global attention module focuses on interactions that span across all positions and channels. This aspect draws inspiration from the non-local operation\cite{nonlocalwang}, which was previously harnessed in models like Dual Attention Network (DANet)\cite{danetfu}. In essence, GLAM combines the strengths of both local and global attention mechanisms to enhance our network.

Furthermore, a thorough explanation of GLAM and its properties can be found in the appendix section of this paper.

\subsection{Cost function}

In a binary classification problem, an imbalance between classes occurs when the number of samples in one class, typically called the minority class, is substantially lower than in the other class, known as the majority class. In many applications, the minority group corresponds to the class of interest, such as the positive class \cite{JohnsonKhoshgoftaar2019}. 
In our study, class imbalance pertains to the distribution of samples between the changed and unchanged areas, where the changed areas constitute the minority group. As mentioned previously, this class imbalance could potentially lead to a bias in the trained model towards the unchanged areas, resulting in the generation of fake data by simply copying and pasting the $S_1T_1$ data as the $S_1T_2$ data instead of learning the true underlying patterns of land cover changes from the Optical Data.
There are three main categories of approaches for addressing class imbalance in machine learning: data-level techniques, algorithm-level methods, and hybrid approaches. Data-level techniques aim to reduce class imbalance through various sampling methods. Algorithm-level methods involve modifying the learning algorithm or its output, often through the use of weight or cost schemes, to reduce bias towards the majority group. Hybrid approaches combine both sampling and algorithmic methods in a strategic manner to address class imbalance \cite{krawczyk2016learningfromimbalance}.
Algorithm-level techniques aim to adapt the learning algorithms to mitigate bias towards the majority group. This requires a deep understanding of the modified algorithm and precise identification of the reasons for its failure in handling skewed distributions. A widely used approach in this context is cost-sensitive learning. \cite{zhou2010multiclasscost}, where the model is modified to assign varying penalties to each group of examples. By attributing greater weight to the underrepresented group, we increase its significance throughout the learning phase, with the objective of reducing the overall cost associated with errors. For example, focal loss \cite{lin2017focalloss} uses the probability of ground truth classes to scale their loss in order to balance the training.
However, in this study, adding a workflow to calculate and use a hard classified change map could potentially add complexity to our already complex model, so we took a much simpler approach. A new loss function called change weighted L1 Loss (CWL1) was proposed to improve the ability of a model to focus on small and scarce changed areas in a scene without using a hard classification map. The proposed loss function utilizes weighted mean to calculate the mean absolute error (L1) over a given area\cite{wang2011infoweighting}.
\begin{equation}
\operatorname{WL1} = \sum_{i=1}^{n} \frac{\sum_{i=1}^{n} w_i \left| y_i - \hat{y}_i \right|}{\sum_{i=1}^{n} w_i}
\label{WL1}
\end{equation}

To calculate the total cost function, the weights of each pixel were determined based on the change map, where $\hat{y}_i$ represents the true value and $y_i$ represents the predicted value for each pixel, and $w_i$ represents the weight for each pixel.

Two weight maps were used in this study, namely the change weight map (CWM) and the reversed change weight map (RCWM). The CWM was calculated as the absolute difference between $S_1T_1$ and $S_1T_2$, with values ranging from $0$ to $1$. On the other hand, the RCWM was obtained as 
Equation\ref{eq:rcwm}:

\begin{equation}
\operatorname{RCWM} = \max\{\operatorname{CWM}\} - \operatorname{CWM} + \min\{\operatorname{CWM}\}
\label{eq:rcwm}
\end{equation}

The total cost function was then calculated as Equation\ref{totalcost}, where $RCWL1$ represents the $WL1$ loss with RCWM as the weight map and $CWL1$ represents the $WL1$ loss with CWM as the weight map.

\begin{equation}
Total\_Cost = \frac{\operatorname{RCWL1} + \gamma \times \operatorname{CWL1}}{1 + \gamma}
\label{totalcost}
\end{equation}

To ensure optimal model performance, the hyperparameter $\gamma$ must be carefully chosen, taking into account the scarcity and size of the changes present in the dataset. Higher values of $\gamma$ can result in the model prioritizing the changed areas and generating less accurate outputs for unchanged regions. Furthermore, in datasets with scarce changes, using a low value of $\gamma$ may cause the model to get biased towards copying the $S_1T_1$ output as the $S_1T_2$ output. Therefore, the selection of an appropriate value of $\gamma$ is essential for achieving optimal results in the data Temporal Shifting task.

\subsection{Evaluation Metrics}
In order to assess the performance of our model, we employed two widely used metrics in the context of data-to-data translation: the structural similarity Iindex (SSIM) \cite{zhou2004imagequality} and the peak signal-to-noise ratio (PSNR). However, to facilitate a more comprehensive evaluation of our results, we also introduced modified versions of these metrics, termed Weighted mean structural similarity index (WSSIM) and weighted peak signal-to-noise ratio (WPSNR). These weighted metrics take into account the importance of each pixel in the assessment process through the incorporation of a weight map. In the following subsections, we first elucidate how SSIM and PSNR function and then detail how these metrics are modified to incorporate a weight map in the evaluation process.

\subsubsection{WSSIM}

SSIM is calculated as Equation\ref{ssim}

\begin{equation}
\operatorname{SSIM}(\mathbf{x}, \mathbf{y})=\frac{\left(2 \mu_x \mu_y+C_1\right)\left(2 \sigma_{x y}+C_2\right)}{\left(\mu_x^2+\mu_y^2+C_1\right)\left(\sigma_x^2+\sigma_y^2+C_2\right)}
\label{ssim}
\end{equation}

Where the standard deviation of the simulated values and real values are denoted by $\sigma_x$ and $\sigma_y$, respectively, while $\mu_x$ and $\mu_y$ represent the mean of the simulated and real values. The covariance between the real and simulated values is denoted by $\sigma_{xy}$. In addition, $C_1$ and $C_2$ are constants introduced to improve the stability of SSIM\cite{ke2013ssim}.

Nonetheless, it is useful to apply SSIM locally rather than globally. Wang et al. \cite{ke2013ssim} used an $11 \times 11$ circular-symmetric Gaussian weighting function $\mathbf{w} = \left\{w_i \mid i=1,2,\ldots,N\right\}$, with a standard deviation of 1.5 samples, normalized to unit sum ($\sum_{i=1}^{N} w_i = 1$). Then, the $\mu_x$, $\sigma_x$, and $\sigma_{xy}$ can be modified as follows:

\begin{equation}
\mu_x=\sum_{i=1}^N w_i x_i
\label{mux}
\end{equation}
\begin{equation}
\sigma_x = \left(\sum_{i=1}^{N} w_i \left(x_i - \mu_x\right)^2\right)^{\frac{1}{2}}
\label{sigmax}
\end{equation}
\begin{equation}
\sigma_{xy} = \sum_{i=1}^{N} w_i \left(x_i - \mu_x\right) \left(y_i - \mu_y\right)
\label{sigmaxy}
\end{equation}

To implement the change weighing factor, using the Change Weight Map captured by the \(j\)th local window, \(\mathbf{\kappa}_j = \left\{\kappa_{j,i} \mid i=1,2,\ldots,N\right\}\), where \(\kappa_{j,i}\) is the CWM weight at each pixel in the \(j\)th local window:

\begin{equation}
\mathrm{K}_j=\sum_{i=1}^N w_i \kappa_{j, i}
\label{kaj}
\end{equation}

Then, to obtain a single quality measure that can assess the overall quality of the data. we utilized the WSSIM to evaluate the overall data quality.

\begin{equation}
\operatorname{WSSIM}(\mathbf{X}, \mathbf{Y})=\frac{\sum_{j=1}^M K_j \operatorname{SSIM}\left(\mathbf{x}_j, \mathbf{y}_j\right)}{\sum_{j=1}^M \mathrm{~K}_j}
\label{wsim}
\end{equation}

\begin{figure}[!t]
\centering
\includegraphics[width=3in]{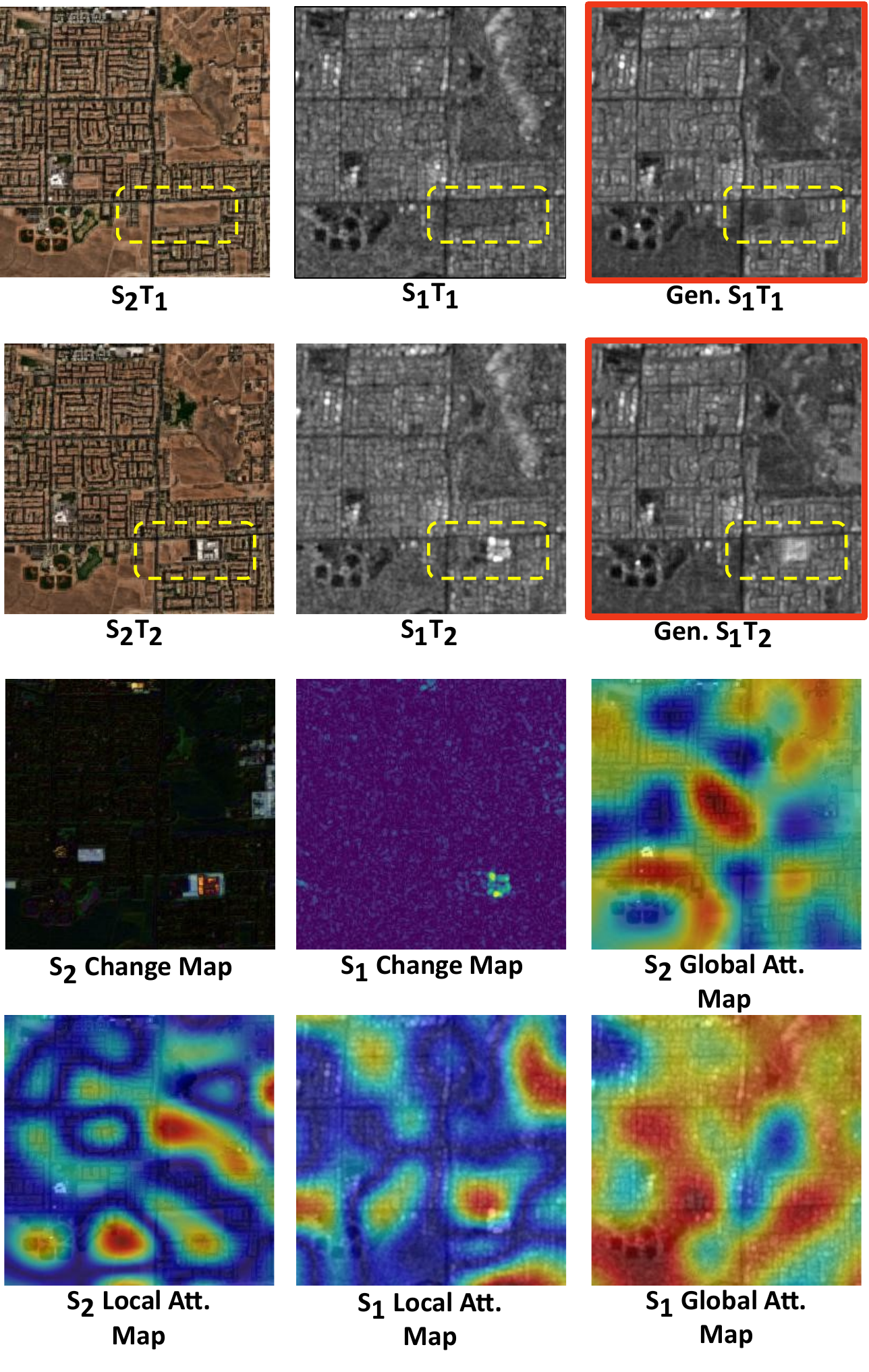}
\caption{Example of the network output. $S_2$ Change Map is a false color composite where Red is the max change in RGB values, Green is the change in NIR values, and Blue is the max change in SWIR values.}
\label{fig:tsgan-v3-output}
\end{figure}

where \(\mathbf{X}\) and \(\mathbf{Y}\) are the reference and the generated data, respectively. \(\mathbf{x}_j\) and \(\mathbf{y}_j\) are the data contents at the \(j\)th local window, and \(M\) is the number of local windows of the data. It is worth mentioning that with a constant Change map \(WSSIM=MSSIM\). The value range of SSIM is from \(-1\) to \(1\). The closer it is to \(1\), the better the synthesized data is.

\subsubsection{WPSNR}

The peak signal-to-noise ratio (PSNR) is a traditional data quality assessment (IQA) index. Generally, the higher the quality of an image, the higher its PSNR value. The formula can be defined as follows:

\begin{equation}
\operatorname{PSNR} = 10 \log_{10}\left(\frac{R^2}{\operatorname{MSE}}\right)
\label{eq:psnr}
\end{equation}
\begin{equation}
\operatorname{MSE} = \frac{\sum_{i=1}^{N} (y_i - \hat{y}_i)^2}{N}
\label{eq:mse}
\end{equation}

Where R is the maximum possible pixel value of the data.
We simply define weighted PSNR (WPSNR), where instead of MSE, we calculate the WMSE using a weight map\cite{wang2011infoweighting}.

\begin{equation}
\operatorname{WPSNR}=10 \lg \left(\frac{R^2}{\operatorname{WMSE}}\right)
\label{eq:wpsnr}
\end{equation}

\begin{equation}
\operatorname{WMSE}=\frac{\sum_{i=1}^N w_i\left(y_i-\hat{y}_i\right)^2}{N \sum_{i=1}^n\left(w_i\right)}
\label{eq:wmse}
\end{equation}

Where \(w_i\) is the weight of each pixel in the weight map.

\section{Ablation experiments setup}\label{sec:ablation}
\subsection{\textbf{Evaluation}}
In this section, we outline the experiments employed to evaluate our model, incorporating both spatial and temporal dimensions. For the spatial dimension, our evaluation encompasses both changed and unchanged areas. Meanwhile, to address the temporal dimension, we assess the model's performance in generating data from both the past and the future.

\subsubsection{\textbf{Spatial evaluation}}
While we have previously elucidated the utilization of a soft change map to weight the loss function, it is important to note that the same approach cannot be applied to the model evaluation. The rationale behind this lies in the inherent limitations of a fuzzy change map, which does not distinctly delineate the model's performance on changed and unchanged regions. This ambiguity arises due to the non-zero weight of changed pixels in the calculation of metrics for unchanged areas and vice versa.

Consequently, to ensure a robust evaluation, a deliberate selection process was undertaken. Specifically, we extracted 154 patches from our test dataset, each featuring discernible urban changes. For these patches, a  thresholding methodology, followed by morphological operations, was employed to create hard binary change maps. These resulting binary maps provided a clear demarcation, enabling the separate evaluation of model performance on both changed and unchanged regions. This approach ensures a more precise and insightful assessment of our models.

\subsubsection{\textbf{Temporal evaluation}}
Urban areas exhibit a tendency to expand over time, resulting in the transformation of bare land or green spaces into developed structures. Acknowledging this phenomenon, our model's evaluation encompasses two distinct scenarios: Backward Temporal Shifting (BTS) and Forward Temporal Shifting (FTS).

In the first scenario (BTS), where $T2<T1$, we input SAR data from 2021 and anticipate the model to generate data from 2019  with the aid of Optical data input from 2019. Consequently, the model is mostly tasked with removing buildings and generating open spaces or green areas.

Conversely, the second scenario (FTS), where $T2>T1$, involves inputting SAR data from 2019 and expecting the model to produce data resembling those from 2021. Here, the model is challenged to convert undeveloped regions into buildings, roads, and similar urban elements.

\begin{figure}[!b]
\centering
\includegraphics[width=3in]{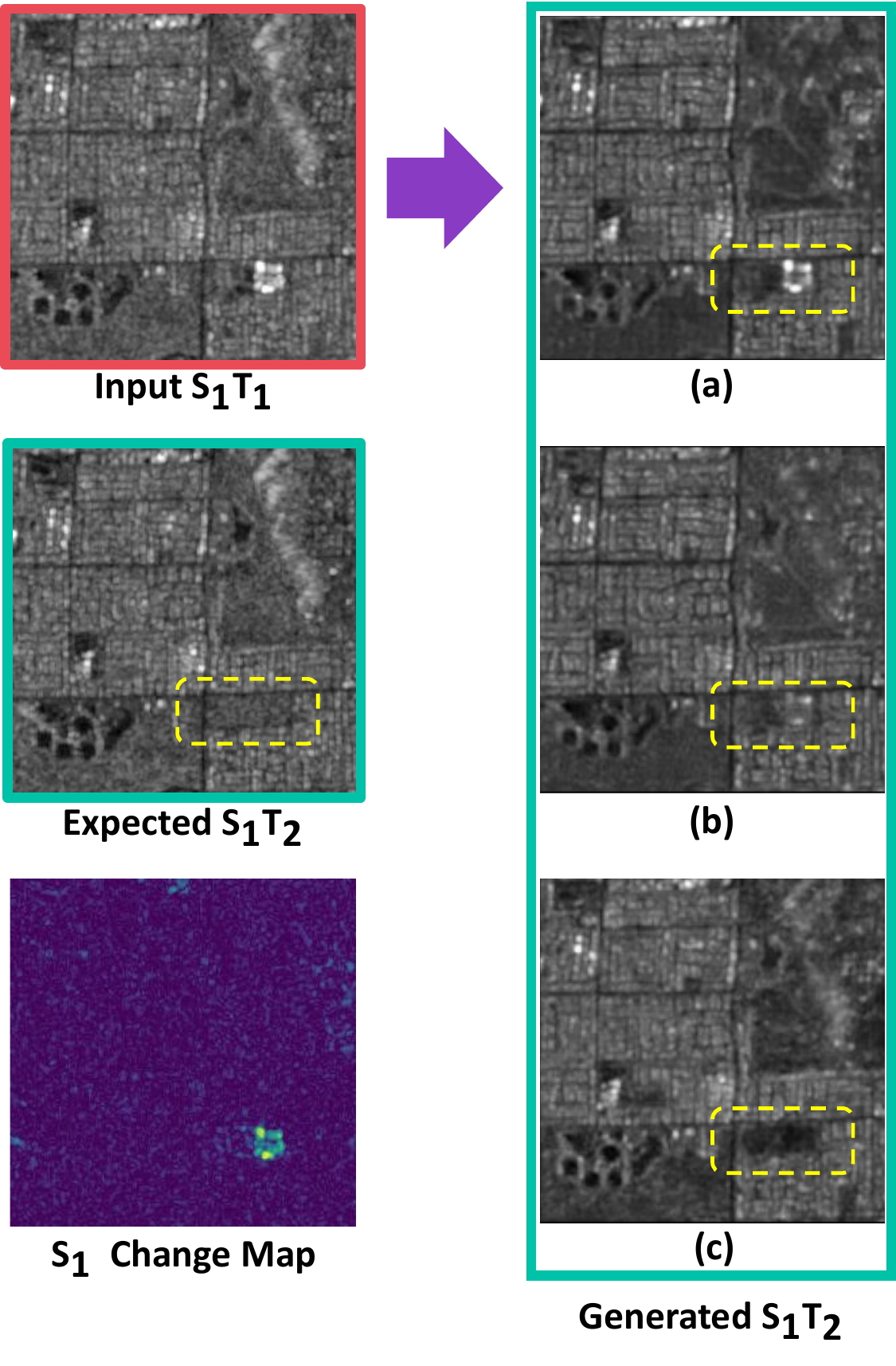}
\caption{Effect of using $\operatorname{CWL1}$ loss and change maps as input: (a) Generated $S_1T_2$ with normal L1 and no input change maps, (b) Generated $S_1T_2$ only using CWL1, (c) Generated $S_1T_2$ using both CWL1 and $S_2$ change map as input.}
\label{fig:losseffect}
\end{figure}

\subsection{\textbf{Models}}\label{sec:models}

In our study, we conducted an ablation analysis on the proposed model, specifically focusing on removing its attention mechanisms, and then compared it with the Pix2Pix model.

We evaluated three versions of our TSGAN model:
\begin{enumerate}
\item \textbf{TSGAN V3: }This version incorporates both the GLAM and SE attention mechanisms, as was described in the methodology.

\item \textbf{TSGAN V2:} In this version, the GLAM module is deactivated, and TSGAN only utilizes the SE mechanism within the fusion component.

 \item \textbf{TSGAN V1:} This is the base model without any attention mechanisms.
\end{enumerate}

Subsequently, we compared the performance of these models with the Pix2Pix model, which was trained under two different scenarios.

\begin{enumerate}
\setcounter{enumi}{3}  

\item \textbf{Original Pix2Pix:} In this setting, the model is the same as the original Pix2Pix, focusing solely on translating optical data into their corresponding SAR data. This scenario does not involve temporal shifting, as it does not use SAR data from a different time as an input, and the model solely learns a translation between optical and SAR data for a specific time. We included this setting to underscore the importance of time-shifting methods compared to simple translation models.

\begin{figure}[!t]
\centering
\includegraphics[width=3in]{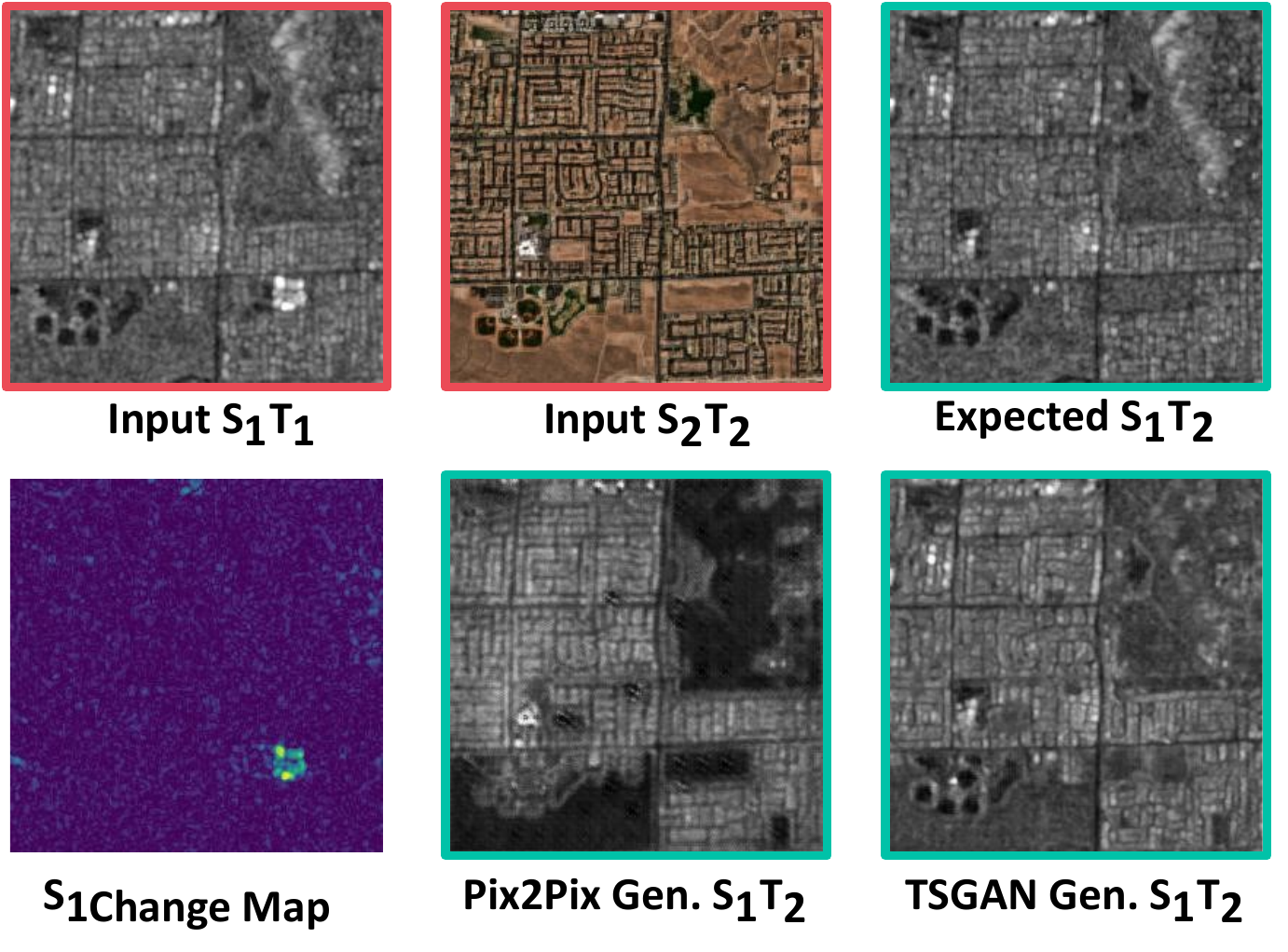}
\caption{Comparison between Temporal Shifting (TSGAN) and traditional translation (Pix2Pix). TSGAN preserves unchanged areas and introduces minor changes in altered regions, whereas Pix2Pix relies on fictional modifications throughout the data.}
\label{fig:tsganvspix2pix}
\end{figure}

\item \textbf{Dual encoder Pix2Pix:} To ensure a fair comparison, we modified the Pix2Pix architecture by duplicating the encoder part and making it a siamese encoder. This modification allowed us to train the model with the same setup as our TSGAN model, enabling independent input of S1 and S2 data. We refer to this modified Pix2Pix model as DE-Pix2Pix. \end{enumerate}

This comprehensive evaluation enabled us to determine the contributions and effectiveness of different attention mechanisms in our model compared to the Pix2Pix architecture. The detailed results of these assessments can be found in Table~\ref{Results}.

\subsection{\textbf{Loss and S2 change map input}}\label{sec:losschange}
In addition to the input $S_1T_1$ and $S_2T_2$ data, we incorporated the changes in the optical data between $T1$ and $T2$, derived as $S_2T_2 - S_1T_1$. This change map, represented as $S_2 Change Map$, is stacked on the optical data input with the objective of providing the model with additional contextual information about the areas in the SAR data that require modification. Additionally, a reversed version of the $S_2 Change Map$ was overlaid on the SAR data to serve a similar purpose in the SAR branch of the model.
To assess the effect of the CWL1 cost function and $S_2 Change Maps$ as input, we conducted tests on the base model under three different settings. The results are presented in Figure~\ref{fig:losseffect} and Table\ref{Results}.

\begin{enumerate}
    \item \textbf{Typical L1 loss}: The model was trained using the standard L1 loss. The input features comprised solely the $S_1T_1$ and
    $S_2T_2$ components, excluding the $S_2 Change Maps$.
    
    \item \textbf{CWL1 integration}: In the second configuration, the CWL1 (change-weighted L1) loss function was introduced. However, similar to the first setting, change maps were not included in the input.
    
    \item \textbf{CWL1 with change maps}: The final setup involved the utilization of both CWL1 and $S_2 Change Maps$ as inputs.
\end{enumerate}

It is important to mention that all of the above losses were accompanied by the Discriminator's loss.

\begin{table*}[!h]
\renewcommand{\arraystretch}{1.3}
\caption{Performance Results of Model Modifications and Other Methods. CWL1 represents the use of the weighted L1 loss function, and S2-CM represents the utilization of $S_2 Change Map$ as input.}
\label{Results}
\centering
\begin{tabular}{c c c c c | c c c c c c}
\hline
\textbf{Model} & \textbf{CWL1} & \textbf{S2-CM} & \textbf{SE} & \textbf{GLAM} & \textbf{PSNR}$\uparrow$ & \textbf{UC-PSNR}$\uparrow$ & \textbf{C-PSNR}$\uparrow$ & \textbf{SSIM}$\uparrow$ & \textbf{UC-SSIM}$\uparrow$ & \textbf{C-SSIM}$\uparrow$ \\
\hline
\multicolumn{11}{c}{\textbf{Whole dataset}} \\

TSGAN V3*& \checkmark& \checkmark & \checkmark & \checkmark & 21.60 & 22.13 & \textbf{16.07} & 0.586 & 0.597 & 0.379 \\

TSGAN V2& \checkmark& \checkmark & \checkmark & \ding{55} & \textbf{21.72} & 22.27 & 16.05 & 0.588 & 0.599 & 0.377 \\

TSGAN V1& \checkmark& \checkmark & \ding{55} & \ding{55} & 21.69 & \textbf{22.28} & 15.87 & 0.598 & 0.601 & \textbf{0.380}\\

TSGAN V1& \checkmark& \ding{55} & \ding{55} & \ding{55} & 21.22 & 21.78 & 15.56 & 0.577 & 0.588 & 0.368\\

TSGAN V1& \ding{55}& \ding{55} & \ding{55} & \ding{55} & 21.41 & 22.09 & 15.12 & \textbf{0.617} & \textbf{0.629} & 0.374\\

DE-Pix2Pix& \checkmark& \checkmark & \ding{55} & \ding{55} & 21.02 & 21.58 & 15.38 & 0.551 & 0.562 & 0.336 \\

Pix2Pix& \ding{55}& \ding{55} & \ding{55} & \ding{55} & 17.57 & 17.73 & 15.31 & 0.311 & 0.315 & 0.235 \\

\hline
\multicolumn{11}{c}{\textbf{Forward Temporal Shifting: $T2>T1$}} \\

TSGAN V3*& \checkmark& \checkmark & \checkmark & \checkmark & 21.45 & 22.03 & \textbf{15.66} & 0.585 & 0.596 & \textbf{0.370}\\

TSGAN V2& \checkmark& \checkmark & \checkmark & \ding{55} & \textbf{21.53} & \textbf{22.13} & 15.58 & 0.586 & 0.597 & 0.368\\

TSGAN V1& \checkmark& \checkmark & \ding{55} & \ding{55} & 21.48& 22.12 & 15.36 & 0.595 & 0.607 & 0.368\\

TSGAN V1& \checkmark& \ding{55} & \ding{55} & \ding{55} & 21.14 & 21.72 & 15.33 & 0.576 & 0.587 & 0.363\\

TSGAN V1& \ding{55}& \ding{55} & \ding{55} & \ding{55} & 21.27 & 21.98 & 14.81 & \textbf{0.615} & \textbf{0.628} & 0.367\\

DE-Pix2Pix& \checkmark& \checkmark & \ding{55} & \ding{55} & 20.85 & 21.44 & 15.02& 0.549 & 0.561 & 0.329\\

Pix2Pix& \ding{55}& \ding{55} & \ding{55} & \ding{55} & 17.49 & 17.67 & 14.84 & 0.308 & 0.312 & 0.225\\

\hline
\multicolumn{11}{c}{\textbf{Backward Temporal Shifting: $T2<T1$}} \\

TSGAN V3*& \checkmark& \checkmark & \checkmark & \checkmark & 21.78 & 22.28 & 16.48 & 0.590 & 0.599 & 0.388\\

TSGAN V2& \checkmark& \checkmark & \checkmark & \ding{55} & \textbf{21.95} & 22.45 & \textbf{16.53} & 0.591 & 0.601 & 0.385\\

TSGAN V1& \checkmark& \checkmark & \ding{55} & \ding{55} & \textbf{21.95} & \textbf{22.49} & 16.40 & 0.601& 0.611 & \textbf{0.392}\\

TSGAN V1& \checkmark& \ding{55} & \ding{55} & \ding{55} & 21.35 & 21.88 & 15.79 & 0.580 & 0.590 & 0.375\\

TSGAN V1& \ding{55}& \ding{55} & \ding{55} & \ding{55} & 21.60 & 22.24 & 15.43 & \textbf{0.619 }& \textbf{0.631} & 0.382\\

DE-Pix2Pix& \checkmark& \checkmark & \ding{55} & \ding{55} & 21.23 & 21.75 & 15.75 & 0.554 & 0.564 & 0.344\\

Pix2Pix& \ding{55}& \ding{55} & \ding{55} & \ding{55} & 17.69 & 17.81 & 15.80 & 0.315 & 0.318 & 0.244\\

\end{tabular}
\end{table*}

\subsection{\textbf{Training}}
Our training strategy was meticulously designed to facilitate the development of a model capable of Shifting SAR data to both the past and future. This was achieved by constructing a data pipeline that involved inputting $S_{1}^{2019}$, $S_{2}^{2021}$, and $S_{2}CM$ together during one training instance, with an expectation for $S_{1}^{2021}$ as the output. In a complementary instance, the model was fed $S_{1}^{2021}$, $S_{2}^{2019}$, and $S_{2}CM$, with an anticipated output of $S_{1}^{2019}$. This two-way training scheme enabled the model to acquire a balanced training experience, preventing it from favoring either the FTS or BTS tasks. Consequently, the model exhibited enhanced generalization capability across both tasks.

The training of our models was conducted utilizing Tesla P100 GPUs with 16GB of VRAM. The training process spanned 10 epochs for dual encoder models and 15 epochs for the original Pix2Pix model, a duration during which the models demonstrated the most favorable equilibrium in generating outputs with the least amount of overfitting. We chose a learning rate of $1 \times 10^{-4}$
with a batch size of 4. 

\begin{figure*}[t]
\centering
\includegraphics[width=7in]{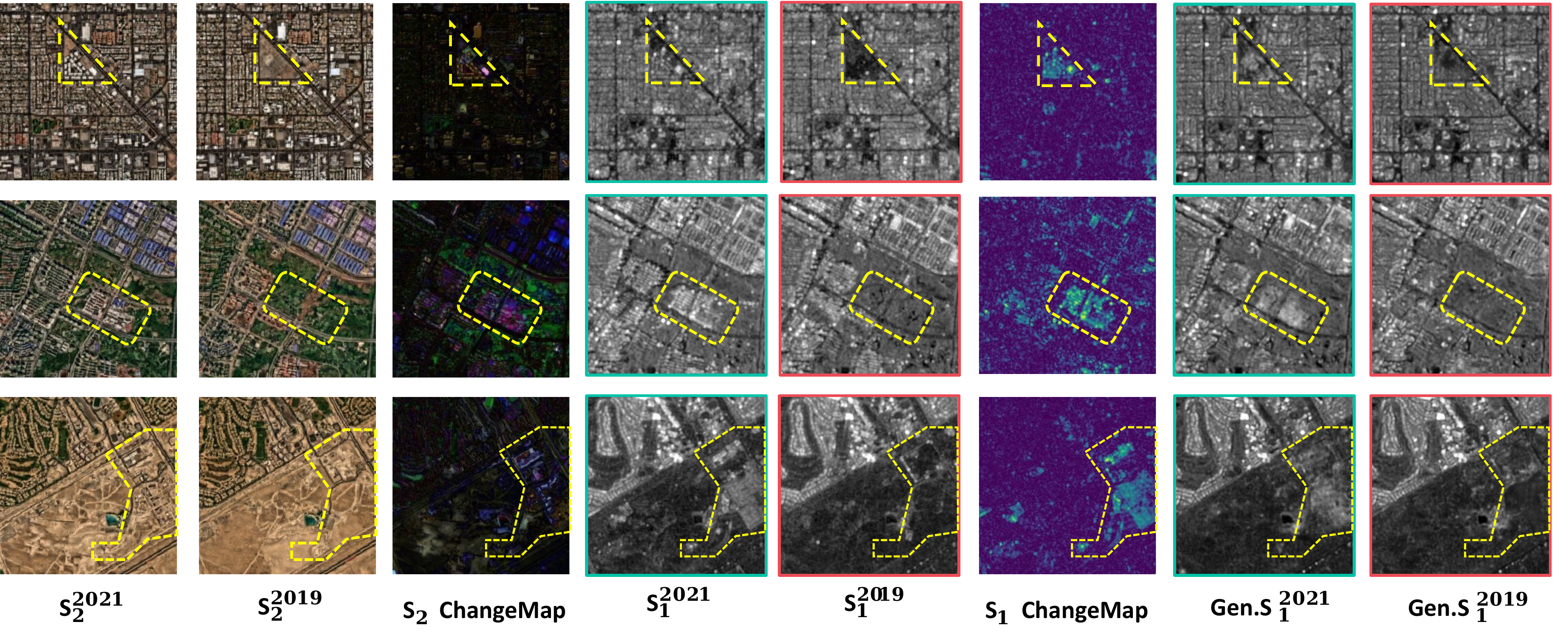}
\caption{More Examples. S2 Change Map False Color Composite: R: $\operatorname{RGB}^{max}$, G: NIR, B: $\operatorname{SWIR}_{1,2}^{max}$ }
\label{fig:more_examples}
\end{figure*}

\section{Results and Discussion}

Table \ref{Results} exhibits the efficacy of our model under varying conditions. As delineated in the ablation experiments section \ref{sec:ablation}, our model was evaluated to answer three distinct questions: 1) how temporal shifting surpasses data translation, 2) how our architecture improves upon the current literature, and 3) why using a change-weighted loss is essential. The responses to these questions will be discussed in the subsections below.

\subsection{Temporal shifting versus translation}
A comparative analysis between Pix2Pix and DE-Pix2Pix reveals the superiority of temporal shifting over conventional translation in unaltered regions while also improving in generating changed regions. Both UC-PSNR and UC-SSIM values demonstrate an improvement, indicating that the model has learned to regenerate unchanged areas from the input SAR data rather than relying solely on translation from optical data. In areas that underwent changes, the performance of temporal shifting also excels, with most metrics showing better results compared to conventional translation, except for C-PSNR in the BTS task. This superior performance in changed regions can be ascribed to the relative ease of modifying SAR data for these areas compared to relying exclusively on optical data.

\subsection{Comparison of TSGAN and Pix2Pix}
A direct comparison between DE-Pix2Pix and TSGAN, using both the CWL1 cost function and $S_2 Change Map$ as input, reveals an immediate improvement in UC-PSNR and UC-SSIM. This indicates that the replacement of the downsampling layer with a 1:1 conv layer on the first skip connection provides a better conduit to transfer unchanged regions into the generated data. The marginal enhancement of the C-SSIM and C-PSNR might be attributed to the larger kernel size of the S2 input skip connection, offering more regional information regarding the translation of changed areas.

\subsection{Impact of change weighted loss and input S2 change map}
Evaluation of TSGAN-V1 under three different settings reveals that without CWL1 and $S_2 Change Map$, it exhibits the highest UC-SSIM and the lowest C-PSNR, confirming the model's tendency to overfit the input SAR data. This behavior resembles a copy-and-paste operation. Additionally, as depicted in Figure\ref{fig:losseffect}, this model struggles to remove corner reflector hot spots compared to other settings. Introducing CWL1 without the input, S2-CM, results in a slight improvement in C-PSNR but a decrease in UC-PSNR, indicating the model's confusion in drawing information in unchanged regions.

\subsection{Effect of attention}
TSGAN-V3, incorporating both GLAM and SE attention modules, demonstrates the highest C-PSNR and C-SSIM in the FTS task, indicating superior performance in generating buildings compared to all other models. TSGAN-V2, when compared to TSGAN-V1, exhibits higher C-PSNR and outperforms all other models in the BTS task.

Figure\ref{fig:tsgan-v3-output}  illustrates the output of TSGAN-V3, showcasing results in both the FTS and BTS phases. Visual inspection of the test dataset reveals that the attention maps tend to highlight areas where the S2-CM indicates change. In the FTS phase, generated buildings are more compact and closely resemble actual structures in SAR data, as perceived by a human observer.

We acknowledge that the GLAM global module occasionally focuses on a single point in the input map, which occurs randomly across different training seeds, resulting in significantly lower performance. To address this issue, we manually excluded runs exhibiting this phenomenon, although this may affect the model's reliability. We recommend the use of TSGAN-V2 or careful manual inspection when employing TSGAN-V3.

\subsection{Further discussion and limitations}
Figure\ref{fig:more_examples} shows more outputs from TSGAN-V3; these visuals, complemented by Table\ref{Results}, demonstrate that TSGAN-V3 excels in removing buildings and generating flat and vegetated areas when compared to the FTS task. On the other hand, in the FTS phase, we can observe the discussed fiction phenomenon in the generated buildings. However, due to the input SAR data, unchanged built areas tend to keep their structure without any artifacts, a problem common to all areas in a traditional Opt2SAR translation model. Figure \ref{fig:tsganvspix2pix} demonstrates the difference between Temporal shifting and Traditional Translation approaches. The output of the Pix2Pix model resembles that it merely turns the Optical data into gray-scale data, analogous to the colorization of SAR data\cite{ji2021colorsar,schmitt2018sarcolorizing} without changing the structure of the data. This distinction highlights the superiority of the Temporal Shift Strategy employed by TSGAN-V3.

While our model showed promising results in an urban setting, we argue that our dataset creation workflow, especially its temporal despecking of SAR images, can be challenging in rapidly changing environments, such as fluctuating riversides or seasonal vegetation coverage. This limitation also affects the model's ability to accurately represent natural environments, as it may not capture subtle changes like plant phenology or moisture variations over time. Additionally, moving objects, like ships in harbors, create many bright spots as they move, which are not visible in the optical image and can mislead the model. These factors should be considered in future applications of our dataset creation workflow. We believe that this problem can be addressed in future studies by using mono-temporal speckle filters.

\section{Conclusion}
Building on the insights gained from our evaluation, we argue that the proposed novel approach, encapsulated in the TSGAN model and complemented with a bi-temporal SAR/optical dataset, represents a clear advancement in the field of Opt2SAR data translation. Our initiative to harness the temporal dimension of SAR data for model input significantly mitigates the prevailing problem of fiction that undermines traditional Opt2SAR translation. Furthermore, we have enhanced the Pix2Pix architecture by modifying layer configurations and incorporating attention mechanisms, thereby achieving notable improvements in both regenerating unchanged areas and generating more realistic built-up areas.

Our work opens up several avenues for future research. First, we suggest exploring the use of our proposed CW-SSIM and CW-PSNR metrics as cost functions for training GAN models, as they may further enhance the quality of the generated data. Second, we recommend using higher-resolution optical data and Digital Surface Models to perform Temporal Shifting of SAR data, as this may enable our model to capture more subtle changes in optical data, and elevation anomalies that affect the SAR backscatter values. Third, we acknowledge the limitations of our research due to the usage-time and storage constraints of free GPU computation and online storage services, which prevented us from using a larger dataset. We hope that future studies can overcome these challenges and validate our model on more diverse and complex datasets.



%

\appendices
\section{GLAM Structure}

In the following section, we will discuss how each attention mechanism works in GLAM.

\subsection{Local Attention}
Firstly, the feature tensor $F \in \mathbb{R}^{c \times h \times w}$ is given to a 2D average pooling layer to flatten the tensor into $F" \in \mathbb{R}^{c \times 1 \times 1}$. Then, $F"$ goes through a 1D convolution layer with kernel size $k$ and a sigmoid non-linearity layer to construct the attention map $A_c^l \in \mathbb{R}_+^{c \times 1 \times 1}$. By embedding $A_c^l$ into the feature map $F$, using Equation 1, we attain the feature map with local channel attention $F_c^l \in \mathbb{R}^{c \times h \times w}$.

\begin{equation}
F_c^l = F \odot A_c^l + F
\end{equation}

Rather than utilizing $F_c^l$ as the local spatial attention's input, the original tensor $F$ is fed into a $1 \times 1$ 2D convolution layer to produce a new feature tensor $F'$ with shape $c' \times h \times w$. Subsequently, the local spatial attention is captured at three different scales using three $3 \times 3$ convolution layers with dilation parameters of 1, 3, and 5, yielding kernels with sizes $3 \times 3$, $5 \times 5$, and $7 \times 7$ respectively. The resulting tensors, having the same shape as $F'$, are concatenated with $F$ to form a new tensor with shape $4c' \times h \times w$. Finally, the local spatial attention map $A_s^l$ is obtained using a $1 \times 1$ convolution layer that reduces the channel dimension to 1. The final Local Attention map $F^l$ can be computed using Equation 2.

\begin{equation}
F^l = F_c^l \odot A_s^l + F_c^l
\end{equation}
 
\subsection{Global Attention}
In GLAM's global attention, the Key $(K)$, Query $(Q)$, and Value $(V)$ tensors are required to compute an attention feature map, following the approach of self-attention in transformers~\cite{vaswani2017attention}. In the context of channel attention, $(K_c)$ and $(Q_c)$ are derived from the original feature map $(F)$ by applying two streams of a 1D convolution with a kernel size of $(k)$, followed by a sigmoid function, resulting in $K_c$ and $Q_c$ where $K_c, Q_c \in \mathbb{R}_+^{1 \times c}$. Then, the global channel attention map $A_c^g \in \mathbb{R}_+^{c \times c}$ is calculated by applying the SoftMax function to the outer product of $K_c$ and $Q_c$.

\begin{equation}
A_c^g = \text{softmax}\left(K_c^\top Q_c\right)
\end{equation}

Next, the Value $(V_c)$ is derived by reshaping $F$ to $\mathbb{R}^{hw \times c}$, so the final channel attention map, after reshaping to the original shape, is:

\begin{equation}
G_c = \text{{reshape}}_{c\times h\times w}(V_c A_c^g), \quad G_c \in \mathbb{R}^{c\times h\times w}
\end{equation}

Hence, the global channel attention feature map can be derived as:

\begin{equation}
F_c^g = F \odot G_c
\end{equation}

To derive the global spatial attention feature map, $V_s$, $K_s$, and $Q_s$ are derived using $1\times1$ convolution layers reducing the channel dimension to $c'$, then reshaping, where $V_s$, $K_s$, and $Q_s \in \mathbb{R}^{c' \times hw}$. The attention map can be derived as:

\begin{equation}
A_s^g = \text{softmax}\left(K_s^\top Q_s\right), \quad A_s^g \in \mathbb{R}_+^{hw \times hw}
\end{equation}

The global spatial attention map $(G_s)$ is derived by reshaping the outer product of $V_s$ and $A_s^g$ to the original size.
\begin{equation}
G_s = \text{{reshape}}_{c\times h\times w}(V_s A_s^g), \quad G_s \in \mathbb{R}^{c\times h\times w}
\end{equation}

The Global attention feature map is then obtained as:

\begin{equation}
F^g = F_c^g \odot G_s + F_c^g
\end{equation}

GLAM incorporates the attention feature maps with the original feature map using a weighted average of the global and local attention maps, along with the original feature map. This is done using three learnable scalar parameters, $w_g$, $w_l$, and $w$, which are obtained via a SoftMax function.

\begin{equation}
F^{gl} = w_gF^g + w_lF^l + wF
\end{equation}

\section*{Acknowledgment}

The first author would like to express profound gratitude to Professor Franz Meyer, whose insightful course on Microwave Remote Sensing provided the foundational knowledge necessary for this research. His generosity in sharing course materials played a pivotal role in the development of this paper.

\section*{Declaration of competing interest}

The authors declare that they have no known competing financial interests or personal relationships that could have appeared to influence the work reported in this paper.

\ifCLASSOPTIONcaptionsoff
  \newpage
\fi



%

\bibliographystyle{IEEEtran}
\bibliography{bare_jrnl}

\begin{thebibliography}{10}
\providecommand{\url}[1]{#1}
\csname url@samestyle\endcsname
\providecommand{\newblock}{\relax}
\providecommand{\bibinfo}[2]{#2}
\providecommand{\BIBentrySTDinterwordspacing}{\spaceskip=0pt\relax}
\providecommand{\BIBentryALTinterwordstretchfactor}{4}
\providecommand{\BIBentryALTinterwordspacing}{\spaceskip=\fontdimen2\font plus
\BIBentryALTinterwordstretchfactor\fontdimen3\font minus \fontdimen4\font\relax}
\providecommand{\BIBforeignlanguage}[2]{{%
\expandafter\ifx\csname l@#1\endcsname\relax
\typeout{** WARNING: IEEEtran.bst: No hyphenation pattern has been}%
\typeout{** loaded for the language `#1'. Using the pattern for}%
\typeout{** the default language instead.}%
\else
\language=\csname l@#1\endcsname
\fi
#2}}
\providecommand{\BIBdecl}{\relax}
\BIBdecl

\bibitem{FAN2023OilSemantic}
J.~Fan and C.~Liu, ``Multitask gans for oil spill classification and semantic segmentation based on sar images,'' \emph{IEEE Journal of Selected Topics in Applied Earth Observations and Remote Sensing}, vol.~16, pp. 2532--2546, 2023.

\bibitem{ZHU2023superres}
\BIBentryALTinterwordspacing
F.~Zhu, C.~Wang, B.~Zhu, C.~Sun, and C.~Qi, ``An improved generative adversarial networks for remote sensing image super-resolution reconstruction via multi-scale residual block,'' \emph{The Egyptian Journal of Remote Sensing and Space Science}, vol.~26, no.~1, pp. 151--160, 2023. [Online]. Available: \url{https://www.sciencedirect.com/science/article/pii/S111098232200120X}
\BIBentrySTDinterwordspacing

\bibitem{zhao2022texttoimage}
R.~Zhao and Z.~Shi, ``Text-to-remote-sensing-image generation with structured generative adversarial networks,'' \emph{IEEE Geoscience and Remote Sensing Letters}, vol.~19, pp. 1--5, 2022.

\bibitem{JOZDANI2022ganreview}
\BIBentryALTinterwordspacing
S.~Jozdani, D.~Chen, D.~Pouliot, and B.~{Alan Johnson}, ``A review and meta-analysis of generative adversarial networks and their applications in remote sensing,'' \emph{International Journal of Applied Earth Observation and Geoinformation}, vol. 108, p. 102734, 2022. [Online]. Available: \url{https://www.sciencedirect.com/science/article/pii/S0303243422000605}
\BIBentrySTDinterwordspacing

\bibitem{zhu2017cyclegan}
J.-Y. Zhu, T.~Park, P.~Isola, and A.~A. Efros, ``Unpaired image-to-image translation using cycle-consistent adversarial networks,'' in \emph{Proceedings of the IEEE international conference on computer vision}, 2017, Conference Proceedings, pp. 2223--2232.

\bibitem{chen2021attentive}
H.~Chen, R.~Chen, and N.~Li, ``Attentive generative adversarial network for removing thin cloud from a single remote sensing image,'' \emph{IET Image Processing}, vol.~15, no.~4, pp. 856--867, 2021.

\bibitem{li2020thincloud}
J.~Li, Z.~Wu, Z.~Hu, J.~Zhang, M.~Li, L.~Mo, and M.~Molinier, ``Thin cloud removal in optical remote sensing images based on generative adversarial networks and physical model of cloud distortion,'' \emph{ISPRS Journal of Photogrammetry and Remote Sensing}, vol. 166, pp. 373--389, 2020.

\bibitem{xiong2021noncloud}
Q.~Xiong, L.~Di, Q.~Feng, D.~Liu, W.~Liu, X.~Zan, L.~Zhang, D.~Zhu, Z.~Liu, X.~Yao, and X.~Zhang, ``Deriving non-cloud contaminated sentinel-2 images with rgb and near-infrared bands from sentinel-1 images based on a conditional generative adversarial network,'' \emph{Remote Sensing}, vol.~13, no.~8, 2021.

\bibitem{zhao2022comprehensivegan}
Y.~Zhao, T.~Celik, N.~Liu, and H.-C. Li, ``A comparative analysis of gan-based methods for sar-to-optical image translation,'' \emph{IEEE Geoscience and Remote Sensing Letters}, vol.~19, pp. 1--5, 2022.

\bibitem{fu2021reciprocal}
S.~Fu, F.~Xu, and Y.-Q. Jin, ``Reciprocal translation between sar and optical remote sensing images with cascaded-residual adversarial networks,'' \emph{Science China Information Sciences}, vol.~64, pp. 1--15, 2021.

\bibitem{doi2020sar2opticcolorregion}
K.~Doi, K.~Sakurada, M.~Onishi, and A.~Iwasaki, ``Gan-based sar-to-optical image translation with region information,'' in \emph{IGARSS 2020 - 2020 IEEE International Geoscience and Remote Sensing Symposium}, 2020, pp. 2069--2072.

\bibitem{YANG2022improvedcgancolorloss}
\BIBentryALTinterwordspacing
X.~Yang, J.~Zhao, Z.~Wei, N.~Wang, and X.~Gao, ``Sar-to-optical image translation based on improved cgan,'' \emph{Pattern Recognition}, vol. 121, p. 108208, 2022. [Online]. Available: \url{https://www.sciencedirect.com/science/article/pii/S0031320321003897}
\BIBentrySTDinterwordspacing

\bibitem{reyes2019sar2optic}
M.~Fuentes~Reyes, S.~Auer, N.~Merkle, C.~Henry, and M.~Schmitt, ``Sar-to-optical image translation based on conditional generative adversarial networks—optimization, opportunities and limits,'' \emph{Remote Sensing}, vol.~11, no.~17, 2019.

\bibitem{xiong2021cloudmultitemp}
\BIBentryALTinterwordspacing
Q.~Xiong, L.~Di, Q.~Feng, D.~Liu, W.~Liu, X.~Zan, L.~Zhang, D.~Zhu, Z.~Liu, X.~Yao, and X.~Zhang, ``Deriving non-cloud contaminated sentinel-2 images with rgb and near-infrared bands from sentinel-1 images based on a conditional generative adversarial network,'' \emph{Remote Sensing}, vol.~13, no.~8, 2021. [Online]. Available: \url{https://www.mdpi.com/2072-4292/13/8/1512}
\BIBentrySTDinterwordspacing

\bibitem{he2021stanfordsuperres}
Y.~He, D.~Wang, N.~Lai, W.~Zhang, C.~Meng, M.~Burke, D.~Lobell, and S.~Ermon, ``Spatial-temporal super-resolution of satellite imagery via conditional pixel synthesis,'' \emph{Advances in Neural Information Processing Systems}, vol.~34, pp. 27\,903--27\,915, 2021.

\bibitem{JohnsonKhoshgoftaar2019}
J.~M. Johnson and T.~M. Khoshgoftaar, ``Survey on deep learning with class imbalance,'' \emph{Journal of Big Data}, vol.~6, no.~1, 2019.

\bibitem{mirza2014cgan}
M.~Mirza and S.~Osindero, ``Conditional generative adversarial nets,'' \emph{arXiv preprint arXiv:1411.1784}, 2014.

\bibitem{isola2017pix2pix}
P.~Isola, J.-Y. Zhu, T.~Zhou, and A.~A. Efros, ``Image-to-image translation with conditional adversarial networks,'' in \emph{Proceedings of the IEEE conference on computer vision and pattern recognition}, 2017, pp. 1125--1134.

\bibitem{wang2019cycle}
L.~Wang, X.~Xu, Y.~Yu, R.~Yang, R.~Gui, Z.~Xu, and F.~Pu, ``Sar-to-optical image translation using supervised cycle-consistent adversarial networks,'' \emph{IEEE Access}, vol.~7, pp. 129\,136--129\,149, 2019.

\bibitem{zhu2017BicycleGAN}
J.-Y. Zhu, R.~Zhang, D.~Pathak, T.~Darrell, A.~A. Efros, O.~Wang, and E.~Shechtman, ``Toward multimodal image-to-image translation,'' \emph{Advances in neural information processing systems}, vol.~30, 2017.

\bibitem{radford2015cut}
A.~Radford, L.~Metz, and S.~Chintala, ``Unsupervised representation learning with deep convolutional generative adversarial networks,'' \emph{arXiv preprint arXiv:1511.06434}, 2015.

\bibitem{huang2018munit}
X.~Huang, M.-Y. Liu, S.~Belongie, and J.~Kautz, ``Multimodal unsupervised image-to-image translation,'' in \emph{Proceedings of the European conference on computer vision (ECCV)}, 2018, Conference Proceedings, pp. 172--189.

\bibitem{chen2020reusingnicegan}
R.~Chen, W.~Huang, B.~Huang, F.~Sun, and B.~Fang, ``Reusing discriminators for encoding: Towards unsupervised image-to-image translation,'' in \emph{Proceedings of the IEEE/CVF conference on computer vision and pattern recognition}, 2020, pp. 8168--8177.

\bibitem{lin2021attentionattncyclegan}
Y.~Lin, Y.~Wang, Y.~Li, Y.~Gao, Z.~Wang, and L.~Khan, ``Attention-based spatial guidance for image-to-image translation,'' in \emph{Proceedings of the IEEE/CVF Winter Conference on Applications of Computer Vision}, 2021, pp. 816--825.

\bibitem{schmitt2018sen12}
M.~Schmitt, L.~H. Hughes, and X.~X. Zhu, ``The sen1-2 dataset for deep learning in sar-optical data fusion,'' \emph{arXiv preprint arXiv:1807.01569}, 2018.

\bibitem{Zhou208unetpp}
Z.~Zhou, M.~M. Rahman~Siddiquee, N.~Tajbakhsh, and J.~Liang, ``Unet++: A nested u-net architecture for medical image segmentation,'' in \emph{Deep Learning in Medical Image Analysis and Multimodal Learning for Clinical Decision Support: 4th International Workshop, DLMIA 2018, and 8th International Workshop, ML-CDS 2018, Held in Conjunction with MICCAI 2018, Granada, Spain, September 20, 2018, Proceedings 4}.\hskip 1em plus 0.5em minus 0.4em\relax Springer, 2018, Conference Proceedings, pp. 3--11.

\bibitem{Li2021deeptranslationchangedetection}
X.~Li, Z.~Du, Y.~Huang, and Z.~Tan, ``A deep translation (gan) based change detection network for optical and sar remote sensing images,'' \emph{ISPRS Journal of Photogrammetry and Remote Sensing}, vol. 179, pp. 14--34, 2021.

\bibitem{hu2023wildfire}
X.~Hu, P.~Zhang, Y.~Ban, and M.~Rahnemoonfar, ``Gan-based sar and optical image translation for wildfire impact assessment using multi-source remote sensing data,'' \emph{Remote Sensing of Environment}, vol. 289, 2023.

\bibitem{gorelick2017gee}
\BIBentryALTinterwordspacing
N.~Gorelick, M.~Hancher, M.~Dixon, S.~Ilyushchenko, D.~Thau, and R.~Moore, ``Google earth engine: Planetary-scale geospatial analysis for everyone,'' \emph{Remote Sensing of Environment}, 2017. [Online]. Available: \url{https://doi.org/10.1016/j.rse.2017.06.031}
\BIBentrySTDinterwordspacing

\bibitem{Wu2020geemap}
\BIBentryALTinterwordspacing
Q.~Wu, ``geemap: A python package for interactive mapping with google earth engine,'' \emph{Journal of Open Source Software}, vol.~5, no.~51, p. 2305, 2020. [Online]. Available: \url{https://doi.org/10.21105/joss.02305}
\BIBentrySTDinterwordspacing

\bibitem{raiyani2021scl}
\BIBentryALTinterwordspacing
K.~Raiyani, T.~Gonçalves, L.~Rato, P.~Salgueiro, and J.~R. Marques~da Silva, ``Sentinel-2 image scene classification: A comparison between sen2cor and a machine learning approach,'' \emph{Remote Sensing}, vol.~13, no.~2, 2021. [Online]. Available: \url{https://www.mdpi.com/2072-4292/13/2/300}
\BIBentrySTDinterwordspacing

\bibitem{duguay202150yearlake}
\BIBentryALTinterwordspacing
J.~Murfitt and C.~R. Duguay, ``50 years of lake ice research from active microwave remote sensing: Progress and prospects,'' \emph{Remote Sensing of Environment}, vol. 264, p. 112616, 2021. [Online]. Available: \url{https://www.sciencedirect.com/science/article/pii/S0034425721003369}
\BIBentrySTDinterwordspacing

\bibitem{woodhouse2017introduction}
I.~H. Woodhouse, ``Geometric distortions in radar images,'' in \emph{Introduction to Microwave Remote Sensing}.\hskip 1em plus 0.5em minus 0.4em\relax CRC Press, 2017, pp. 281--284.

\bibitem{daudt2018oscd}
R.~C. Daudt, B.~L. Saux, A.~Boulch, and Y.~Gousseau, ``Urban change detection for multispectral earth observation using convolutional neural networks,'' 2018.

\bibitem{munoz2020temporal}
A.~Munoz, M.~Zolfaghari, M.~Argus, and T.~Brox, ``Temporal shift gan for large scale video generation,'' 2020.

\bibitem{donahue2019labelconditioned}
D.~Donahue, ``Label-conditioned next-frame video generation with neural flows,'' 2019.

\bibitem{urbangrowthrate}
\BIBentryALTinterwordspacing
M.~Zhao, C.~Cheng, Y.~Zhou, X.~Li, S.~Shen, and C.~Song, ``A global dataset of annual urban extents (1992--2020) from harmonized nighttime lights,'' \emph{Earth System Science Data}, vol.~14, no.~2, pp. 517--534, 2022. [Online]. Available: \url{https://essd.copernicus.org/articles/14/517/2022/}
\BIBentrySTDinterwordspacing

\bibitem{hafner2022urban}
S.~Hafner, Y.~Ban, and A.~Nascetti, ``Urban change detection using a dual-task siamese network and semi-supervised learning,'' in \emph{IGARSS 2022-2022 IEEE International Geoscience and Remote Sensing Symposium}.\hskip 1em plus 0.5em minus 0.4em\relax IEEE, 2022, pp. 1071--1074.

\bibitem{vaswani2017attention}
A.~Vaswani, N.~Shazeer, N.~Parmar, J.~Uszkoreit, L.~Jones, A.~N. Gomez, {\L}.~Kaiser, and I.~Polosukhin, ``Attention is all you need,'' \emph{Advances in neural information processing systems}, vol.~30, 2017.

\bibitem{Dosovitvisiontransformer}
A.~Dosovitskiy, L.~Beyer, A.~Kolesnikov, D.~Weissenborn, X.~Zhai, T.~Unterthiner, M.~Dehghani, M.~Minderer, G.~Heigold, and S.~Gelly, ``An image is worth 16x16 words: Transformers for image recognition at scale,'' \emph{arXiv preprint arXiv:2010.11929}, 2020.

\bibitem{tokenvisiontransformer}
L.~Yuan, Y.~Chen, T.~Wang, W.~Yu, Y.~Shi, Z.-H. Jiang, F.~E. Tay, J.~Feng, and S.~Yan, ``Tokens-to-token vit: Training vision transformers from scratch on imagenet,'' in \emph{Proceedings of the IEEE/CVF international conference on computer vision}, 2021, Conference Proceedings, pp. 558--567.

\bibitem{boutosvitsar}
N.~I. Bountos, D.~Michail, and I.~Papoutsis, ``Learning class prototypes from synthetic insar with vision transformers,'' \emph{arXiv preprint arXiv:2201.03016}, 2022.

\bibitem{chenvitcd}
H.~Chen, Z.~Qi, and Z.~Shi, ``Remote sensing image change detection with transformers,'' \emph{IEEE Transactions on Geoscience and Remote Sensing}, vol.~60, pp. 1--14, 2021.

\bibitem{pangvitcdsar}
L.~Pang, J.~Sun, Y.~Chi, Y.~Yang, F.~Zhang, and L.~Zhang, ``Cd-transunet: A hybrid transformer network for the change detection of urban buildings using l-band sar images,'' \emph{Sustainability}, vol.~14, no.~16, 2022.

\bibitem{oktay2018attentionunet}
O.~Oktay, J.~Schlemper, L.~L. Folgoc, M.~Lee, M.~Heinrich, K.~Misawa, K.~Mori, S.~McDonagh, N.~Y. Hammerla, and B.~Kainz, ``Attention u-net: Learning where to look for the pancreas,'' \emph{arXiv preprint arXiv:1804.03999}, 2018.

\bibitem{khanh2020enhanceunetatteniton}
\BIBentryALTinterwordspacing
T.~L.~B. Khanh, D.-P. Dao, N.-H. Ho, H.-J. Yang, E.-T. Baek, G.~Lee, S.-H. Kim, and S.~B. Yoo, ``Enhancing u-net with spatial-channel attention gate for abnormal tissue segmentation in medical imaging,'' \emph{Applied Sciences}, vol.~10, no.~17, 2020. [Online]. Available: \url{https://www.mdpi.com/2076-3417/10/17/5729}
\BIBentrySTDinterwordspacing

\bibitem{zhao2020scanet}
\BIBentryALTinterwordspacing
P.~Zhao, J.~Zhang, W.~Fang, and S.~Deng, ``Scau-net: Spatial-channel attention u-net for gland segmentation,'' \emph{Frontiers in Bioengineering and Biotechnology}, vol.~8, 2020. [Online]. Available: \url{https://www.frontiersin.org/articles/10.3389/fbioe.2020.00670}
\BIBentrySTDinterwordspacing

\bibitem{HuShenSun2018SqueezeExcitation}
J.~Hu, L.~Shen, and G.~Sun, ``Squeeze-and-excitation networks,'' in \emph{Proceedings of the IEEE Conference on Computer Vision and Pattern Recognition}, 2018, Conference Proceedings, pp. 7132--7141.

\bibitem{rangzan2022stripe}
M.~{Rangzan} and S.~{Attarchi}, ``{Removing Stripe Noise from Satellite Images using Convolutional Neural Networks in Frequency Domain},'' in \emph{EGU General Assembly Conference Abstracts}, ser. EGU General Assembly Conference Abstracts, May 2022, pp. EGU22--12\,575.

\bibitem{SongHanAvrithis2023GLAM}
C.~H. Song, H.~J. Han, and Y.~Avrithis, ``All the attention you need: Global-local, spatial-channel attention for image retrieval,'' in \emph{Proceedings of the IEEE/CVF Winter Conference on Applications of Computer Vision}, 2023, Conference Proceedings, pp. 2754--2763.

\bibitem{cbamwoo}
S.~Woo, J.~Park, J.-Y. Lee, and I.~S. Kweon, ``Cbam: Convolutional block attention module,'' in \emph{Proceedings of the European conference on computer vision (ECCV)}, 2018, Conference Proceedings, pp. 3--19.

\bibitem{nonlocalwang}
X.~Wang, R.~Girshick, A.~Gupta, and K.~He, ``Non-local neural networks,'' in \emph{Proceedings of the IEEE conference on computer vision and pattern recognition}, 2018, Conference Proceedings, pp. 7794--7803.

\bibitem{danetfu}
J.~Fu, J.~Liu, H.~Tian, Y.~Li, Y.~Bao, Z.~Fang, and H.~Lu, ``Dual attention network for scene segmentation,'' in \emph{Proceedings of the IEEE/CVF conference on computer vision and pattern recognition}, 2019, Conference Proceedings, pp. 3146--3154.

\bibitem{krawczyk2016learningfromimbalance}
B.~Krawczyk, ``Learning from imbalanced data: open challenges and future directions,'' \emph{Progress in Artificial Intelligence}, vol.~5, no.~4, pp. 221--232, 2016.

\bibitem{zhou2010multiclasscost}
Z.~Zhou and X.~Liu, ``On multi‐class cost‐sensitive learning,'' \emph{Computational Intelligence}, vol.~26, no.~3, pp. 232--257, 2010.

\bibitem{lin2017focalloss}
T.-Y. Lin, P.~Goyal, R.~Girshick, K.~He, and P.~Dollár, ``Focal loss for dense object detection,'' in \emph{Proceedings of the IEEE international conference on computer vision}, 2017, Conference Proceedings, pp. 2980--2988.

\bibitem{wang2011infoweighting}
Z.~Wang and Q.~Li, ``Information content weighting for perceptual image quality assessment,'' \emph{IEEE Transactions on Image Processing}, vol.~20, no.~5, pp. 1185--1198, 2011.

\bibitem{zhou2004imagequality}
Z.~Wang, A.~Bovik, H.~Sheikh, and E.~Simoncelli, ``Image quality assessment: from error visibility to structural similarity,'' \emph{IEEE Transactions on Image Processing}, vol.~13, no.~4, pp. 600--612, 2004.

\bibitem{ke2013ssim}
K.~Gu, G.~Zhai, X.~Yang, W.~Zhang, and M.~Liu, ``Structural similarity weighting for image quality assessment,'' in \emph{2013 IEEE International Conference on Multimedia and Expo Workshops (ICMEW)}, 2013, pp. 1--6.

\bibitem{ji2021colorsar}
G.~Ji, Z.~Wang, L.~Zhou, Y.~Xia, S.~Zhong, and S.~Gong, ``Sar image colorization using multidomain cycle-consistency generative adversarial network,'' \emph{IEEE Geoscience and Remote Sensing Letters}, vol.~18, no.~2, pp. 296--300, 2021.

\bibitem{schmitt2018sarcolorizing}
\BIBentryALTinterwordspacing
M.~Schmitt, L.~H. Hughes, M.~K\"orner, and X.~X. Zhu, ``Colorizing sentinel-1 sar images using a variational autoencoder conditioned on sentinel-2 imagery,'' \emph{The International Archives of the Photogrammetry, Remote Sensing and Spatial Information Sciences}, vol. XLII-2, pp. 1045--1051, 2018. [Online]. Available: \url{https://isprs-archives.copernicus.org/articles/XLII-2/1045/2018/}
\BIBentrySTDinterwordspacing

\end{thebibliography}

%








\end{document}